%% file: main.tex
\documentclass{article}

\PassOptionsToPackage{compress}{natbib}
\usepackage[preprint]{neurips_2026}

\input{packages}

\input{notations}

\title{
    \memo: Memory as a Model
}

\author{
    Ryan Wei Heng Quek$^{1,2,3,4}$\thanks{Equal contributions and $^\dagger$Corresponding author.}
    \quad Sanghyuk Lee$^{5,6,7\hspace{0.3mm}*}$
    \quad Alfred Wei Lun Leong$^{4,8\hspace{0.3mm}*}$ \\
    \textbf{Arun Verma}$^{9\hspace{0.3mm}*\dagger}$
    \quad \textbf{Alok Prakash}$^{9}$ 
    \quad \textbf{Nancy F. Chen}$^{3}$ \\
    \textbf{Bryan Kian Hsiang Low}$^{1,2,4,9}$
    \quad \textbf{Daniela Rus}$^{7,9}$
    \quad \textbf{Armando Solar-Lezama}$^{7,9}$
    \\
    $^{1}$Institute of Data Science, National University of Singapore, Singapore\\
    $^{2}$Integrative Sciences and Engineering Programme, NUSGS, Singapore\\
    \quad $^{3}$Agency for Science, Technology, Research (A*STAR), Singapore\\
    $^{4}$Department of Computer Science, National University of Singapore, Singapore\\
    $^{5}$University of Tokyo, Japan
    \quad $^{6}$Liquid AI, USA \\ 
    $^{7}$CSAIL, Massachusetts Institute of Technology, USA 
    \quad $^{8}$AI Singapore
    \\
    $^{9}$Singapore-MIT Alliance for Research and Technology Centre, Singapore
    \\
    \texttt{ryanquekweiheng@u.nus.edu}
    \quad \texttt{leesanghyuk@g.ecc.u-tokyo.ac.jp} \\
    \texttt{alfred\_leong@u.nus.edu}
    \quad \texttt{arun.verma@smart.mit.edu} \\
    \texttt{alok.prakash@smart.mit.edu}
    \quad \texttt{nancy\_chen@a-star.edu.sg}\\
    \texttt{lowkh@comp.nus.edu.sg}
    \quad \texttt{rus@csail.mit.edu} 
    \quad \texttt{asolar@csail.mit.edu}
}

\begin{document}

    \maketitle

    \begin{abstract}
        \input{latex/abstract}
    \end{abstract}



    \section{Introduction}
    \label{sec:introduction}
    \input{latex/introduction}

    \section{Related Work}
    \label{sec:related_work}
    \input{latex/related_work}

    \section{Preliminaries}
    \label{sec:problem}
    \input{latex/problem}

    \section{MeMo: Memory as a Model}
    \label{sec:memo}
    \input{latex/memo}

    \section{Experiments}
    \label{sec:experiments}
    \input{latex/experiment}

    \section{Conclusion}
    \label{sec:conclusion}
    \input{latex/conclusion}

    \clearpage
    \bibliographystyle{unsrt}
	\bibliography{references}

    \clearpage
    \appendix
    \input{latex/appendix}

     \hrule height 0.5mm

\end{document}

%% file: packages.tex
\usepackage[utf8]{inputenc} 
\usepackage[T1]{fontenc}    
\usepackage{url}            
\usepackage{booktabs}       
\usepackage{amsfonts}       
\usepackage{nicefrac}       
\usepackage{microtype}      
\usepackage{xcolor}         

\usepackage{graphicx}
\usepackage{rotating}
\usepackage{tabularx}
\usepackage{pdflscape}
\usepackage{amsmath,amsthm,amssymb}
\usepackage{algorithm,algorithmic}
\usepackage{bbm,dsfont}
\usepackage{subfig}
\usepackage{caption}
\usepackage{cancel}
\usepackage{enumerate,cases}
\usepackage{thmtools,thm-restate}
\usepackage{mathtools}

\usepackage{multirow}
\usepackage{makecell}
\usepackage{setspace}
\usepackage{pifont}
\usepackage{array}
\usepackage{enumitem}
\usepackage{lineno}     
\usepackage{tcolorbox}
\tcbuselibrary{skins}
\usepackage{tablefootnote}
\usepackage[dvipsnames]{xcolor}

\allowdisplaybreaks

\usepackage{hyperref}		 
\hypersetup{
	colorlinks	= true,		 
	urlcolor     = blue,	 
	linkcolor	 = purple, 	 
	citecolor    = violet    
}
\usepackage[capitalize]{cleveref}
\crefformat{figure}{Fig.~#2#1#3}       
\crefformat{table}{Tab.~#2#1#3}        
\crefformat{equation}{Eq.~(#2#1#3)}    
\crefformat{section}{Sec.~#2#1#3}      
\crefformat{appendix}{App.~#2#1#3}     
\crefformat{algorithm}{Alg.~#2#1#3}    
\crefformat{definition}{Def.~#2#1#3}       

\usepackage{todonotes}

\usepackage{silence}
\usepackage{placeins}

%% file: notations.tex
\newcommand{\memo}{{\textsc{MeMo}}}     
\newcommand{\gen}{\textsc{Generator} model}
\newcommand{\mem}{\textsc{Memory} model}
\newcommand{\memory}{\textsc{Memory}}
\newcommand{\main}{\textsc{Executive} model}

\newcommand{\para}[1]{\textbf{#1}~}

\newcommand{\Prob}[1]{\bP\left\{#1\right\}}

\newcommand{\R}{\bR}

\renewcommand{\phi}{\varphi}
\renewcommand{\epsilon}{\varepsilon}

\newcommand{\step}[1]{{\large \textcircled{\small #1}}}

\newcommand{\el}{\end{flushleft}}
\newcommand{\bl}{\begin{flushleft}}

\newcommand{\squishlisttwo}{
 \begin{list}{$\bullet$}
  { \setlength{\itemsep}{1pt}
     \setlength{\parsep}{0pt}
    \setlength{\topsep}{0pt}
    \setlength{\partopsep}{0pt}
    \setlength{\leftmargin}{1em}
    \setlength{\labelwidth}{1.5em}
    \setlength{\labelsep}{0.5em} } 
}
\newcommand{\squishend}{\end{list}}

\newcommand{\bP}{\mathbb{P}}

\newcommand{\bR}{\mathbb{R}}

\newcommand{\cD}{\mathcal{D}}

\newcommand{\cG}{\mathcal{G}}

\newcommand{\cK}{\mathcal{K}}

\newcommand{\cM}{\mathcal{M}}

\newcommand{\cQ}{\mathcal{Q}}

\newcommand{\cS}{\mathcal{S}}

\theoremstyle{plain}

\newtheorem{defi}{Definition}

%% file: latex/abstract.tex

Large language models (LLMs) achieve strong performance across a wide range of tasks, but remain frozen after pretraining until subsequent updates. 
Many real-world applications require timely, domain-specific information, motivating the need for efficient mechanisms to incorporate new knowledge.
In this paper, we introduce \textbf{\memo{}} (\textbf{Me}mory as a \textbf{Mo}del), a modular framework that encodes new knowledge into a dedicated \mem{} while keeping the LLM parameters
unchanged. 
Compared to existing methods, \memo{} offers several advantages: 
\textbf{(a)}~it captures complex cross-document relationships, 
\textbf{(b)}~it is robust to retrieval noise, 
\textbf{(c)}~it avoids catastrophic forgetting in the LLM, 
\textbf{(d)}~it does not require access to the LLM's weights or output logits, enabling plug-and-play integration with both open and proprietary closed-source LLMs, and
\textbf{(e)}~its retrieval cost is independent of corpus size at inference time.
Our experimental results on three benchmarks, BrowseComp-Plus, NarrativeQA, and MuSiQue, show that \memo{} achieves strong performance compared to existing methods across diverse settings.

%% file: latex/introduction.tex

Large language models (LLMs) have demonstrated remarkable capabilities across diverse tasks~\citep{kojima2023largelanguagemodelszeroshot, ArXiv23_zhao2023survey, survey-llms-code-generation}. 
Despite their successes, these models are effectively \emph{frozen} for extended periods after pretraining~\citep{xu2024knowledgeconflictsllmssurvey} until subsequent updates, 
causing their pretrained knowledge to become increasingly outdated as the world evolves. 
For applications that require up-to-date~\citep{cheng2024dateddatatracingknowledge, kasai2024realtimeqawhatsanswer} or domain-specific~\citep{singhal2022largelanguagemodelsencode, wu2023bloomberggptlargelanguagemodel} knowledge, this dependence on static knowledge presents a fundamental architectural limitation~\citep{lewis2021retrievalaugmentedgenerationknowledgeintensivenlp, kandpal2023largelanguagemodelsstruggle}.
Retraining is a natural solution but remains prohibitively expensive at modern scales~\citep{wu2022sustainable}, motivating the need for an efficient mechanism to integrate new external knowledge into LLMs without full retraining.

\begin{figure}[!ht] 
    \centering
    \includegraphics[width=\linewidth]{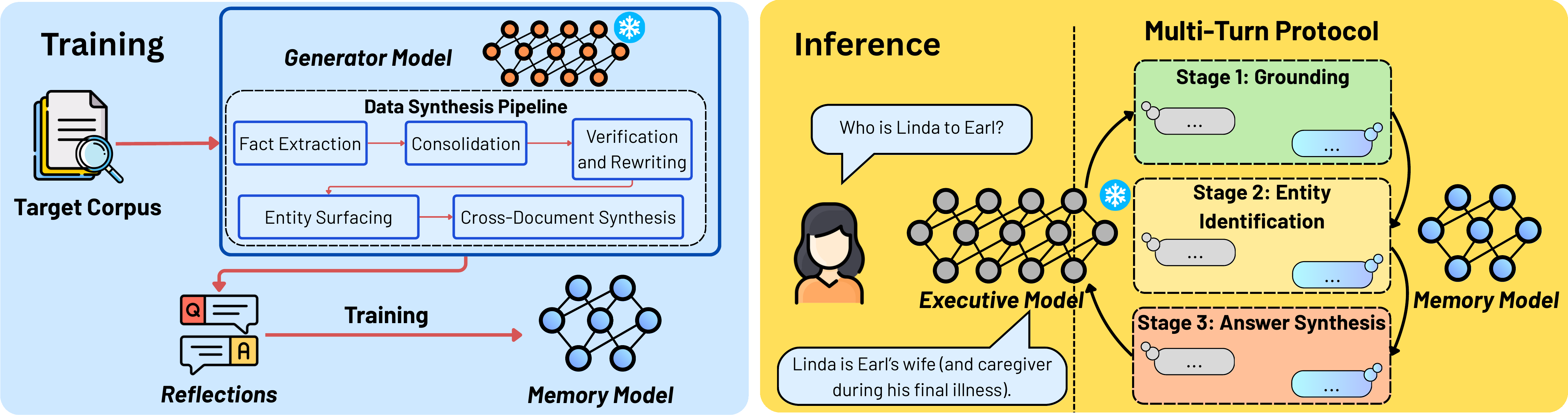}
    \caption{
        \textbf{Overview of the training and inference pipeline 
        of \memo{}.}
        During \mem{} training (left), a frozen \gen{} transforms a target corpus into a reflection QA dataset via fact extraction, consolidation, verification, entity surfacing, and cross-document synthesis, which is then used to train a dedicated \mem{}.
        During inference (right), the frozen \main{} answers complex user queries by querying the \mem{} through a structured multi-turn protocol: it decomposes the input into simpler, targeted sub-queries, retrieves intermediate responses from the \mem{}, and reasons over them to produce a final answer to the user's query.
    }
    \label{fig:MeMo}
    \vspace{-4mm}
\end{figure}

Existing methods for integrating new knowledge into LLMs fall into three categories.
\step{1} \textit{Non-parametric methods} retrieve relevant information from an external store at inference time via lexical~\citep{bm25}, dense~\citep{nvembedv2}, or graph-based retrievers~\citep{lewis2020retrieval,graphrag,gutierrez2024hipporag,gutierrez2025rag}, before incorporating it through in-context learning~\citep{incontextlearning,dong-etal-2024-survey-in-context-learning}. 
However, these methods are constrained by limited context windows and struggle to synthesize cross-document relationships when relevant information is distributed across multiple documents~\citep{tang2024multihop, lin2025optimizingmultihopdocumentretrieval}. 
\step{2} \textit{Parametric methods} internalize knowledge directly into model parameters via continual pretraining~\citep{ke2023continual} or fine-tuning~\citep{ouyang2022training,wang2023self,chung2024scaling} on the target corpus directly. While effective, they are computationally expensive, prone to catastrophic forgetting~\citep{luo2025empiricalstudycatastrophicforgetting}, and tend to memorize training distributions rather than acquire transferable knowledge,
limiting generalization to unseen queries~\citep{chu2025sft}.
\step{3} \textit{Latent memory methods}~\citep{chevalier2023autocompressor, mu2023gist, ge2024icae, zhang2026memgen} compress knowledge into soft tokens or other model-specific representations, but suffer from representation coupling: the memory is tightly bound to the specific model used to produce these representations, limiting transferability across LLMs.

We introduce \textbf{\memo{}} (\textbf{Me}mory as a \textbf{Mo}del), a modular framework where a dedicated \mem{} is trained on new knowledge, and an \main{} retrieves relevant information from the \mem{} at inference time via targeted sub-queries and then reasons over the retrieved information to respond to user queries.
\memo{} combines the complementary strengths of the three paradigms above while mitigating their individual limitations.
Like the non-parametric methods, it is able to leverage off-the-shelf frontier models unchanged by separating the memory from the reasoning model; it shares with the parametric methods the ability to internalize knowledge in model parameters, and it shares the benefits of a compact, queryable memory artifact with latent memory methods.
As a result, \memo{} offers the following advantages: \textbf{(a)} it captures complex cross-document relationships, 
\textbf{(b)} it is robust to retrieval noise, 
\textbf{(c)} it avoids catastrophic forgetting by keeping the \main{} parameters unchanged, 
\textbf{(d)} it does not require access to the \main{}’s weights or output logits, enabling plug-and-play integration with both open and proprietary LLMs, and
\textbf{(e)} its retrieval cost is independent of corpus size at inference time due to the fixed size of the \mem{}.
However, designing \memo{} to comprehensively capture cross-document relationships during training while accurately answering arbitrary queries at inference time introduces two key challenges, which we outline below and address them with novel methods.

\textbf{\step{1} Training \mem{}.}
A core challenge in the \mem{} is ensuring it can accurately answer diverse, unseen queries at inference time, including those requiring cross-document reasoning and long-context understanding.
A natural approach is to train directly on the raw corpus using standard data augmentation techniques such as paraphrasing~\citep{li2022data,chen2023empirical,allen2024physics}, additional sampling of generated QA pairs~\citep{alberti2019synthetic,puri2020training}, or targeted gap-filling, where the model identifies and completes missing knowledge from the corpus~\citep{feng2024don,jie2024self}. However, these approaches fail to consolidate related facts into compositional representations necessary for robust generalization to unseen queries~\citep{chu2025sft}.
With this challenge in mind, we design a novel \emph{five-step data synthesis pipeline} guided by a \gen{} (\cref{sec:datagen}) that distills the corpus into a question–answer (QA) dataset of \emph{reflections}: compositional 
representations that expose underlying corpus knowledge under diverse query variations (illustrated in \cref{fig:MeMo} (left) and details in \cref{sec:datagen}).
We train \mem{} on the synthesized reflection QA dataset via supervised fine-tuning (see \cref{sec:training}), enabling \mem{} to capture more complex, cross-document relationships and compositional structure than retrieval-based methods.

\textbf{\step{2} Querying \mem{}.}
At inference time, complex or compositional queries often require multi-step reasoning and aggregation of information across multiple documents. Naively querying \mem{} via single-turn or unstructured multi-turn interactions fails to reliably retrieve the knowledge required to answer such queries.
To address this, we design a \emph{three-stage inference pipeline} in which \main{} queries and retrieves information from \mem{} via a structured multi-turn protocol, decomposing complex user queries into targeted sub-queries that align with the shared reflection interface (illustrated in \cref{fig:MeMo} (right) and more details are in \cref{sec:inference}). 
Unlike retrieval-based methods, this approach incurs retrieval cost independent of corpus size and is robust to retrieval noise (see~\cref{subsec:noise_ablation}).
Crucially, because \memo{} treats \main{} as a black box and does not access its weights, gradients, or output logits, it supports \emph{plug-and-play} integration with \emph{any LLM}, including both open and proprietary closed-source models.

Our method is guided by a single design principle: \emph{reflections}, corpus-derived structures that require no knowledge of future queries, yet naturally serve as the precise interface through which any query can access the underlying corpus without ever observing it directly. During training, the \mem{} internalizes these reflections; \main{} retrieves relevant knowledge through targeted sub-queries at inference time.
Building on the challenges outlined above and the methods proposed to address them, we summarize the key contributions of this paper as follows:
\vspace{-2mm}
\begin{itemize}[leftmargin=*]
	\setlength\itemsep{0.0em}
    \item \textbf{Novel data synthesis pipeline.} 
    We propose a five-step data synthesis pipeline that uses a \gen{}, an LLM that may be the same as or smaller than \main{}, to distill a target corpus into reflections, enabling a dedicated \mem{} to internalize knowledge in compositional forms that capture more complex cross-document relationships and generalize robustly to diverse, unseen query variations at inference time (see \cref{sec:datagen,sec:training}).

    \item \textbf{Structured multi-turn protocol.}
    We introduce a \emph{structured} multi-turn protocol that systematically decomposes complex queries into targeted sub-queries aligned with the shared reflection interface. The protocol supports plug-and-play integration with any arbitrary LLM, including proprietary closed-source LLMs, and has retrieval cost independent of corpus size (see \cref{sec:inference}).
    
    \item  \textbf{Empirical validation.}
    We evaluate \memo{} on BrowseComp-Plus, NarrativeQA, and MuSiQue, demonstrating strong performance against both parametric and non-parametric baselines. We further empirically validate \memo{}'s robustness to retrieval noise (see \cref{sec:experiments}).  
    
\end{itemize}

%% file: latex/related_work.tex

\para{Non-parametric methods.}
Non-parametric alternatives~\citep{bm25, nvembedv2, gutierrez2025rag} avoid parameter updates entirely, instead, supplying new knowledge at inference time.
In particular, in-context learning (ICL)~\citep{incontextlearning, dong-etal-2024-survey-in-context-learning} inserts relevant knowledge directly into the prompt, avoiding catastrophic forgetting.
However, ICL scales poorly with increasing context length: the computational cost of autoregressive generation~\citep{vaswani2023attentionneed} leads to substantial token overhead and inference latency as the knowledge base grows~\citep{gelada2025scalingcontextrequiresrethinking}, and even explicitly long-context models exhibit significant performance degradation as context length increases~\citep{liu2024lost, hsieh2024ruler}.
Retrieval-augmented generation (RAG)~\citep{lewis2020retrieval,graphrag,gutierrez2024hipporag,gutierrez2025rag} addresses this scalability bottleneck by selectively retrieving relevant chunks of knowledge at inference time.
However, RAG systems are highly sensitive to retrieval noise~\citep{powerofnoise}, where irrelevant or misleading passages substantially degrade generation quality~\citep{liu2026tacklinginherentdifficultynoise,zhang2026understanding}.
In addition, RAG systems often struggle to reason over complex cross-document dependencies~\citep{tang2024multihop}, as they lack robust mechanisms for synthesizing information that is distributed across multiple chunks or a large corpus~\citep{lin2025optimizingmultihopdocumentretrieval}.

\para{Parametric methods.}
Existing post-training approaches, such as continual pretraining on new corpora~\citep{ke2023continual, sun2020ernie-cpt} or supervised fine-tuning (SFT) on curated instruction data~\citep{ouyang2022training,wang2023self,chung2024scaling}, attempt to address this limitation by incorporating new knowledge into LLMs during post-training. 
While conceptually straightforward, these parametric methods often suffer from catastrophic forgetting, whereby adaptation to newly observed knowledge degrades previously acquired knowledge, learned capabilities~\citep{luo2025empiricalstudycatastrophicforgetting,learningwithoutforgetting, harmon2025mappingposttrainingforgettinglanguage}, and safety alignment learned during LLM post-training~\citep{qi2024fine}. 
In addition, the scale of modern LLMs makes frequent fine-tuning computationally expensive~\citep{zhang2023dissecting, xia2024understanding}, and fine-tuning is often infeasible for proprietary, closed-source models~\citep{manchanda2025opensourceadvantagelarge}, substantially limiting the practicality of parametric methods in real-world, large-scale applications.

\para{Latent memory methods.}
Another approach to storing knowledge is via \emph{compressed latent representations}, which lie between non-parametric retrieval and fully parametric methods. Context compression techniques such as AutoCompressor~\citep{chevalier2023autocompressor}, Gist tokens~\citep{mu2023gist}, and ICAE~\citep{ge2024icae} encode knowledge into compact soft tokens prepended at inference, reducing ICL token overhead without discarding information. 
However, these representations are tightly coupled to the encoder and cannot be consumed by other model families, limiting compatibility with black-box LLMs.
Similarly, recurrent-state models~\citep{gu2023mamba, sun2023retnet} and nearest-neighbor memory methods such as Memorizing Transformers~\citep{wu2022memorizing} and $k$NN-LM~\citep{khandelwal2020generalization} rely on model-specific representations or architectures, preventing post hoc use with pretrained LLMs.
Although Memory Decoder~\citep{cao2025memory} is a plug-and-play pretrained memory module that integrates without modifying model parameters, it is limited to architectures sharing a common tokenizer, enabling reuse only within this subset.
The core limitation of these methods is \emph{representation coupling}: latent memory is inseparable from the model that produces it. In contrast, \memo{} allows a plug-and-play integration with any LLM, including closed-source models.

\begin{table}[!ht]
    \vspace{-2mm}
    \centering
    \caption{
        A comparison of desirable properties across different memory paradigms, showing that \memo{} satisfies them through its modular memory construction and memory-augmented reasoning.
    }
    \label{tab:memo_comparison}
    \vspace{1mm}
    \setlength{\tabcolsep}{6pt}
    \renewcommand{\arraystretch}{1.25}
    \resizebox{\textwidth}{!}{%
    \begin{tabular}{|l|c|c|c|c|c|c|}
    \hline \textbf{Methods}
    & \makecell{\textbf{Frozen}\\\textbf{base LLM}}
    & \makecell{\textbf{No}\\\textbf{retrieval index}}
    & \makecell{\textbf{Black-box}\\\textbf{compatible}}
    & \makecell{\textbf{No catastrophic}\\\textbf{forgetting}}
    & \makecell{\textbf{Constant-size}\\\textbf{memory}}
    & \makecell{\textbf{Cross-LLM}\\\textbf{transferable}} \\
    \hline
    \textbf{Non-parametric} {\small(RAG, ICL)}                  & \checkmark & $\times$ & \checkmark & \checkmark & $\times$ & \checkmark\tablefootnote{We assume a fixed, task-agnostic embedding model decoupled from the \main{} for RAG, enabling the retrieval index to be reused across models. For ICL, prompts are assumed to be raw-text-based, free of model-specific formatting.} \\
    \textbf{Parametric} {\small(CPT, SFT)}                      & $\times$ & \checkmark & $\times$ & $\times$ & \checkmark & $\times$ \\
    \textbf{Latent memory} {\small(AutoCompressor, Gist, ICAE)} & \checkmark & \checkmark & $\times$ & \checkmark & $\times$ & $\times$ \\
    \textbf{\memo{} (Ours)}                                        & \checkmark & \checkmark & \checkmark & \checkmark & \checkmark & \checkmark \\
    \hline
    \end{tabular}
    }
\end{table}

%% file: latex/problem.tex

\para{Problem setting.}
Let $\cM_\theta$ denote a large language model with frozen parameters $\theta \in \R^p$, pretrained on a corpus $\cD_{\text{pre}}$. We treat $\cM_\theta$ as a conditional distribution that maps a prompt $x$ to a response $\cM_\theta(x)$, and assume only black-box access; in particular, $\cM_\theta$ may be either a white-box model or a closed-source model accessed via API. Let $\mathcal{D} = \{d_1, \ldots, d_N\}$ denote a target corpus of $N$ documents containing knowledge that $\mathcal{M}_\theta$ cannot reliably recall\footnote{We do not assume $\mathcal{D}$ is disjoint from $\mathcal{D}_{\text{pre}}$, as training data is rarely disclosed by model providers. A document is considered \textit{effectively absent} from $\mathcal{M}_\theta$'s knowledge if the model fails to answer questions grounded in it, either because it never appeared in $\mathcal{D}_{\text{pre}}$ or because the training process was insufficient to retain it. For more information, refer to~\cref{app:EXPR-data-contamination}.}.
Let $\cQ$ be a set of queries, each $q \in \cQ$ associated with a ground-truth answer $a^\star(q)$ and a set of supporting documents $\cS(q) \subseteq \cD$. Note that $\cS(q)$ is a theoretical construct used to characterize query complexity.

\para{Knowledge integration mechanism.}
A \emph{knowledge integration mechanism} is a pair $(\Phi, f)$, where $\Phi$ maps the corpus to a representation $\cK \doteq \Phi(\cD)$ and $f$ combines $\cK$ with $\cM_\theta$ at inference to produce responses $f(\cM_\theta, \cK, q)$. We formalize the goal as follows.
\begin{defi}[Knowledge Integration Problem]
    \label{defi:kip}
    Given a frozen model $\cM_\theta$ and a target corpus $\cD$, find a mechanism $(\Phi, f)$ that maximizes
    $\mathbb{E}_{q \sim \cQ}\bigl[\Prob{f(\cM_\theta, \Phi(\cD), q) = a^\star(q)}\bigr]$ without modifying $\theta$.
\end{defi}

\para{Existing approaches.}
Existing methods differ in their choice of $(\Phi, f)$. \emph{ICL} sets $\cK = \cD$ and $f(\cM_\theta, \cK, q) = \cM_\theta([\cD; q])$, i.e., appending the corpus directly to the prompt. \emph{RAG} constructs $\cK$ as a retrieval index and defines $f$ to retrieve a subset $\hat{\cS} \subseteq \cD$ before passing $[\hat{\cS}; q]$ to $\cM_\theta$. \emph{Fine-tuning} sets $\cK = \emptyset$ and $f = \cM_{\theta'}$, where $\theta'$ is obtained by updating $\theta$ on $\cD$. In contrast, \memo{} defines $\cK$ as the parameters of a small, dedicated \mem{} $\cM_\phi$ with $\phi \ll \theta$, trained on a reflection QA dataset derived from $\cD$, and queried by a frozen \main{} $\cM_\theta$ at inference time. \cref{tab:memo_comparison} summarizes how these paradigms compare across desirable properties.

%% file: latex/memo.tex

\memo{} addresses the knowledge integration problem (Def.~\ref{defi:kip}) through two components: a frozen \emph{model} $\cM_\theta$ (\main{}), which handles reasoning and responds to user queries, and a \emph{\mem{}} $\cM_\phi$, which is trained to encode knowledge in its parameters from a target corpus $\cD$. 
Our pipeline operates in two phases: (i) a \emph{training phase} that constructs \mem{} from $\cD$, and (ii) an \emph{inference phase} in which \main{} queries and retrieves information from \mem{} to answer knowledge-intensive questions (see \cref{sec:datagen,sec:training,sec:inference}).

\subsection{Data Synthesis Pipeline}
\label{sec:datagen}
Given a corpus of documents $\cD$, our objective in the data generation process is to construct a reflection QA dataset $\cQ_{\text{final}}$ that captures both single-document facts and cross-document relationships. This process is driven by a \gen{} $\cM_{\text{gen}}$ and proceeds through five steps, as summarized in \cref{alg:datagen-pipeline} and illustrated in \cref{fig:MeMo}: 
(1) fact extraction from raw documents, 
(2) consolidation of redundant or overlapping information, 
(3) verification and rewriting to ensure correctness and clarity, 
(4) entity surfacing to explicitly represent key entities, and 
(5) cross-document synthesis to integrate evidence across the corpus. Importantly, no document identifiers or watermarks are embedded in the generated QA pairs at any step, preventing \mem{} from exploiting shortcut signals during evaluation.

\vspace{-1.5mm}
\begin{algorithm}[!ht]
\caption{Reflection QA Dataset Generation Pipeline from Target Corpus}
\label{alg:datagen-pipeline}
\begin{algorithmic}[1]
\REQUIRE Corpus $\cD$, generator $\cM_{\text{gen}}$, document groups $\cG = \{G_1, \ldots, G_R\}$ with $G_i \subseteq \cD$
\STATE $\cQ_{\text{final}} \gets \emptyset$
\FORALL{document $d \in \cD$}
  \STATE $C \gets \mathrm{Chunk}(d)$ \hfill $\vartriangleright$ Segment into chunks
  \STATE $\cQ_{\text{ver}}^{d} \gets \emptyset$
  \FORALL{chunk $c \in C$}
      \STATE $\cQ_{\text{dir}},\, \cQ_{\text{indir}} \gets \cM_{\text{gen}}(c)$ \hfill $\vartriangleright$ Step 1: Direct and indirect extraction
      \STATE $\cQ_{\text{raw}} \gets \cQ_{\text{dir}} \cup \cQ_{\text{indir}}$ \hfill $\vartriangleright$ Step 2a: Merge direct and indirect
      \STATE $\cQ_{\text{mrg}} \gets \cM_{\text{gen}}(\cQ_{\text{raw}})$ \hfill $\vartriangleright$ Step 2b: Consolidate related pairs
      \STATE $\cQ_{\text{con}} \gets \cQ_{\text{raw}} \cup \cQ_{\text{mrg}}$ \hfill $\vartriangleright$ Step 2c: Full merge set
      \STATE $\cQ_{\text{ver}} \gets \cM_{\text{gen}}(\cQ_{\text{con}},\, c)$ \hfill $\vartriangleright$ Step 3: Verify self-containment; rewrite or discard
      \STATE $\cQ_{\text{ver}}^{d} \gets \cQ_{\text{ver}}^{d} \cup\, \cQ_{\text{ver}}$
  \ENDFOR
  \STATE $\cQ_{\text{ent}}^{d} \gets \cM_{\text{gen}}(\cQ_{\text{ver}}^{d})$ \hfill $\vartriangleright$ Step 4: Entity-surfacing pairs
  \STATE $\cQ_{\text{final}} \gets \cQ_{\text{final}} \cup\, \cQ_{\text{ver}}^{d} \cup \cQ_{\text{ent}}^{d}$
\ENDFOR
\FORALL{$G_i \in \cG$}
  \STATE $\cQ_{\text{cross}} \gets \cM_{\text{gen}}\!\Bigl(\bigcup_{d \in G_i} \bigl(\cQ_{\text{ver}}^{d} \cup \cQ_{\text{ent}}^{d}\bigr)\Bigr)$ \hfill $\vartriangleright$ Step 5: Cross-document synthesis
  \STATE $\cQ_{\text{final}} \gets \cQ_{\text{final}} \cup\, \cQ_{\text{cross}}$
\ENDFOR
\RETURN $\cQ_{\text{final}}$
\end{algorithmic}
\end{algorithm}

\para{Step 1: Fact extraction.}
Each document $d \in \cD$ is segmented into chunks $C$, where each chunk corresponds either to an entire document or to a contiguous segment of a longer document.
For each chunk, $\cM_{\text{gen}}$ performs two parallel extraction processes: \emph{direct extraction}, which captures explicitly stated facts (producing $\cQ_{\text{dir}}$), and \emph{indirect extraction}, which targets inferred or synthesized information beyond the surface text (producing $\cQ_{\text{indir}}$). 
This dual extraction process ensures that both factual recall and inferential reasoning are represented in the training signal for \mem{}.

\para{Step 2: Consolidation.}
The \gen{} $\cM_{\text{gen}}$ consolidates $\cQ_{\text{dir}} \cup \cQ_{\text{indir}}$ by identifying QA pairs that share a common underlying context (such as entity, time period, or relationship type) and combining them into QA pairs that encompass multiple facts, denoted $\cQ_{\text{mrg}}$. This merging process produces training instances that require integrating multiple facts within the same contextual chunk, going beyond single-fact question answering pairs. The synthesized QA pairs are subsequently unified with the original sets to form the consolidated dataset $\cQ_{\text{con}} = \cQ_{\text{dir}} \cup \cQ_{\text{indir}} \cup \cQ_{\text{mrg}}$. 

\para{Step 3: Verification and rewriting.}
Each QA pair in $\cQ_{\text{con}}$ is evaluated for \emph{self-containment} by $\cM_{\text{gen}}$, i.e., whether it can be 
fully understood and correctly answered in isolation, without access to the source chunk. 
Common failure modes include unresolved pronouns (e.g., ``What did \emph{they} propose?'') and implicit references (e.g., ``As noted in the above table\ldots''). 
Non-self-contained QA pairs are rewritten by $\cM_{\text{gen}}$ using the source chunk $C$ as context; QA pairs that remain ambiguous after rewriting are discarded. This check-and-rewrite procedure yields the verified set $\cQ_{\text{ver}}$, a set of QA pairs that can be used as training examples without access to the source chunk.

\para{Step 4: Entity surfacing.}
For each named entity in $\cQ_{\text{ver}}$, $\cM_{\text{gen}}$ generates a set of entity-surfacing QA pairs in which the question encodes the entity’s attributes and relationships (including connections to other named entities) and the answer reveals its identity. Facts about each entity are aggregated across all QA pairs within the chunk prior to generation, enabling the integration and composition of information from multiple source pairs. Questions are generated at varying levels of complexity, ranging from single-fact to multi-fact queries. These pairs, denoted $\cQ_{\text{ent}}$, aim to mitigate the reversal curse~\citep{berglund2023reversal,allen2023physics32} by training \mem{} to infer entities from indirect or partially specified descriptions. This capability supports the \emph{entity identification turn} at inference time (\cref{sec:inference}).

\para{Step 5: Cross-document synthesis.}
The final step operates over pre-defined document groups $\cG = \{G_1, \ldots, G_R\}$, where chunks within each group $G_i$ are topically related. 
Such groups arise naturally, for example, when a large document is segmented into chunks (forming a single group) or from human-provided labels. 
For each group $G_i$, $\cM_{\text{gen}}$ is provided with both the verified pairs $\cQ_{\text{ver}}^{d}$ and the entity-surfacing pairs $\cQ_{\text{ent}}^{d}$ for all $d \in G_i$ from all member documents and identifies two types of cross-document connections:
\vspace{-2mm}
\begin{itemize}
	\setlength\itemsep{0.0em}
    \item \emph{Converging clues}: multiple documents provide complementary facts about the same entity, which together enable its identification.
    
    \item \emph{Parallel properties}:  different entities across documents share a common attribute or role, enabling comparative and analogical reasoning.
\end{itemize}

Both types yield QA pairs with support size $\cS(q) > 1$ (\cref{sec:problem}), directly targeting the cross-document synthesis objective. The final dataset is $\cQ_{\text{final}} = \cQ_{\text{ver}} \cup \cQ_{\text{ent}} \cup \cQ_{\text{cross}}$, which collectively captures self-contained, entity-centric, and cross-document reflections for training \mem{}.
Ablations of the pipeline design are presented in \cref{app:datagen-pipeline-Ablation}.

\subsection{Training the \mem{}}
\label{sec:training}
Given $\cQ_{\text{final}}$, \mem{} is trained via supervised fine-tuning to map questions directly to answers \emph{without} access to source documents at inference time. 
\mem{} is initialized from a small pretrained language model, substantially smaller than \main{} (e.g., 1.5B vs.\ 32B parameters), and optimized by minimizing the next-token prediction loss over answer tokens only.
$$
    \mathcal{L}(\phi) \;=\; -\!\!\!\sum_{(q_i,\, a_i)\, \in\, \cQ_{\text{final}}}\;\; \sum_{t=1}^{|a_i|} \log \cM_\phi\!\left(a_i^{(t)} \;\middle|\; q_i,\, a_i^{(1:t-1)}\right).
$$
Conditioning only on the question and preceding answer tokens, and never on source documents, forces \mem{} to internalize knowledge \emph{parametrically} rather than rely on copying from retrieved context. 
This constitutes a key distinction from RAG-based readers: at inference time, \mem{} generates answers solely from its internalized parametric knowledge, without access to any external corpus. 
Further details on hyperparameter choices are provided in \cref{app:ALGO-train-settings} and training paradigms (full SFT vs. LoRA) are provided in \cref{app:EXPR-sft-vs-lora}.

\subsection{Continual Knowledge Integration via Model Merging}
A practical desideratum of any knowledge integration system is the ability to incorporate new corpora incrementally without retraining on or rebuilding from all previously ingested sources. For parametric models, integrating new knowledge typically requires retraining on the union of all observed corpora, a cost that grows prohibitively with the number of sources. In contrast, non-parametric systems such as knowledge graphs and vector databases support efficient incremental updates. We explore \emph{model merging}~\citep{yang2024model} as an approach to close this gap for parametric models. Model merging aims to preserve knowledge from multiple sources without requiring joint training on their union, by combining $K$ \mem{} models, each trained independently on a distinct corpus, into a single model.

\para{Continual knowledge integration.}
Let $\{\mathcal{D}_1, \dots, \mathcal{D}_K\}$ be a collection of pairwise disjoint target corpora. For each corpus $\mathcal{D}_i$, we generate a reflection QA dataset $\mathcal{Q}^{(i)}_{\text{final}}$~(\cref{sec:datagen}) and train a corresponding \textsc{Memory} model $\mathcal{M}_{\varphi_i}$ via SFT (\cref{sec:training}), initializing all $K$ models from the same pretrained base $\mathcal{M}_{\varphi_0}$. We define the \emph{task vector} for $\mathcal{D}_i$ as $\tau_i \;{=}\; \varphi_i - \varphi_0$, capturing the parametric shift induced by training on $\mathcal{D}_i$ alone. The merged \textsc{Memory} model is then obtained as 
$$
    \varphi_{\text{merged}} \;{=}\; \mathrm{Merge}(\varphi_0,\, \{\tau_i\}_{i=1}^{K};\, \Theta),
$$    
where $\Theta$ denotes method-specific hyperparameters (e.g., merging coefficients, sparsification densities). We discuss alternative merging methods and their respective limitations in~\cref{app:ALGO-model-training-discussion}.

\subsection{Inference-Time Integration}
\label{sec:inference}
At inference time, \main{} queries and retrieves information from \mem{} through a structured multi-turn protocol, with \main{} treating \mem{} as an external knowledge oracle. 
The pipeline has three sequential stages, each designed to progressively improve the likelihood of producing a correct final answer, as illustrated in \cref{fig:MeMo} (right). 
Each stage utilizes distinct prompts, sampling temperatures and independent budgets to control the number of interactions between \main{} and \mem{}.

\para{Stage 1: Grounding.}
Given a query $q$, \main{} decomposes it into a set of atomic, clue-probing sub-questions $\{q_1', \ldots, q_J'\}$, where each sub-question targets a single identifying constraint in $q$, and $J$ is adaptively determined by \main{}.
The \mem{} answers each sub-question independently, without shared context, producing grounding responses $\{m_1, \ldots, m_J\}$.
These responses draw on \mem{}'s parametric knowledge to provide additional contextual grounding for subsequent interactions in the later stages.

\para{Stage 2: Entity identification.}
Using the grounding responses as context, \main{} iteratively narrows a set of candidate entities by issuing targeted follow-up sub-queries to \mem{} across multiple interactions. 
This process continues until \main{} converges on a single entity $e^\star$ or the stage budget is exhausted. If no candidates are identified, Stage~3 is skipped and \main{} synthesizes a final answer from the grounding responses alone. 
This stage leverages \mem{}'s training on the entity-surfacing QA pairs $\cQ_{\text{ent}}$ (\cref{sec:datagen}). 

\para{Stage 3: Answer seeking and synthesis.}
Conditioned on the identified entity $e^\star$, \main{} queries \mem{} for additional supporting facts through targeted follow-up questions. 
Once sufficient evidence is gathered, or the stage budget is exhausted, \main{} synthesizes the accumulated responses into a final answer:
$$
    \hat{a} \;=\; \cM_\theta\!\bigl(q,\; \{m_j\}_{j=1}^J,\; e^\star,\; m_{\text{seek}}\bigr).
$$

Notably, the \mem{} responses $m_j$ and $m_{\text{seek}}$ are compact natural-language snippets whose lengths are independent of the corpus size, ensuring constant-time inference.
As all interactions with $\cM_\theta$ occur through its input–output interface, \memo{} remains fully compatible with black-box \main{}s, including proprietary APIs, without requiring access to internal parameters. For full implementation details, refer to \cref{app:EXPR-eval-process-and-prompt} and the supplementary materials.

%% file: latex/experiment.tex

\para{Datasets.}
We evaluate \memo{} on three knowledge-intensive benchmarks. \textbf{BrowseComp-Plus}~\citep{chen2025browsecompplusfairtransparentevaluation} is a deep-research benchmark requiring multi-hop, multi-document retrieval and reasoning; we filter non-English instances with LangDetect~\citep{danilak2021langdetect}, sample 300 questions, and pair each question's evidence documents with an equal number of negative documents,\footnote{BrowseComp-Plus and MuSiQue provide annotations of gold (correct), evidence (supporting), and negative (distractor) documents. Gold documents are a subset of the evidence documents.} yielding 3,541 documents in total.
\textbf{NarrativeQA}~\citep{kovcisky2018narrativeqa} tests discourse understanding over long documents such as books and movie scripts; we use 293 questions across 10\footnote{We follow HippoRAG2 and evaluate on 10 such documents from the NarrativeQA validation split (294 questions); one duplicate is removed for consistency.} documents.
\textbf{MuSiQue}~\citep{trivedi2022musiquemultihopquestionssinglehop} requires composing 2--4 reasoning steps across multiple Wikipedia paragraphs; we use 1,000 questions and construct the target corpus following the same procedure as for BrowseComp-Plus, yielding 5,296 documents. Further details are in \cref{app:EXPR-dataset-preparation}; datasets and code are in the supplementary materials.

\para{Baselines.}
We compare \memo{} against four baselines: \textbf{BM25}~\citep{bm25} (lexical retrieval), \textbf{NV-Embed-V2}~\citep{nvembedv2} (dense retrieval), \textbf{HippoRAG2}~\citep{gutierrez2025rag} (graph-based RAG, state-of-the-art), and \textbf{Cartridges}~\citep{eyuboglu2025cartridges} (a trained KV-cache loaded onto \main{} at inference; the closest existing parametric baseline to \memo{}). Newer methods exist~\citep{chevalier2023autocompressor, cao2025memorydecoderpretrainedplugandplay} but typically require white-box access to \main{} and are therefore not directly comparable. We additionally include \textbf{Perfect Retrieval} as an \textit{empirical upper bound}, where \main{} receives exclusively the evidence documents in context~\citep{incontextlearning}. Retrieval baselines use top-$k{=}9$ with adaptive backoff: reducing $k$ progressively until the retrieved context fits \main{}'s context window.

\para{Implementation and evaluation.}
\emph{(a) Data generation.} We use Qwen2.5-32B-Instruct~\citep{qwen2025qwen25technicalreport} as the \gen{}, served via vLLM~\citep{kwon2023efficientmemorymanagementlarge} with YaRN RoPE scaling~\citep{su2024roformer,peng2024yarn} to support a 131K-token context window during long-context generation.
\emph{(b) Training.} We train \mem{}, which is initialized from Qwen2.5-14B-Instruct for 3 epochs with fused AdamW~\citep{loshchilov2017decoupled} and DeepSpeed~2~\citep{rajbhandari2020zero} at learning rate $2{\times}10^{-5}$; full hyperparameters are provided in \cref{app:ALGO-train-settings}.
\emph{(c) Evaluation.} We instantiate \main{} with either Qwen2.5-32B-Instruct or Gemini-3-Flash~\citep{google2025gemini3flash} to evaluate the same trained \mem{} across models of varying reasoning capability; both models have minimal prior knowledge of the evaluation datasets (\cref{app:EXPR-data-contamination}). \main{} queries \mem{} through the multi-turn protocol described in \cref{sec:inference}. We report binary accuracy judged by Gemini-2.5-Flash-Lite~\citep{comanici2025gemini25pushingfrontier} via DeepEval~\citep{Ip_deepeval_2026}, as mean $\pm$ standard deviation over three runs for Qwen2.5-32B-Instruct and a single run for Gemini-3.0-Flash.
\emph{(d) Continual integration.} For the model-merging experiment (\cref{sec:experiments-merging}), we partition NarrativeQA into two pairwise-disjoint subsets ($K{=}2$, with $\sim$640k QA pairs each), SFT a separate Qwen2.5-14B-Instruct \mem{} on each, and sweep six merging methods at three densities (yielding 14 configurations total).

\subsection{Experimental results}
\para{\memo{} achieves strong performance across benchmarks.}
As shown in Table~\ref{tab:main-results}, \memo{} consistently outperforms all baselines on NarrativeQA and MuSiQue across both \main{}s. On NarrativeQA, the most challenging benchmark (\cref{app:EXPR-data-contamination}), \memo{} achieves $26.85\%$ with Qwen2.5-32B-Instruct and $53.58\%$ with Gemini-3-Flash, substantially surpassing all baselines. This is notable: NarrativeQA requires reasoning over long passages with complex connections, where retrieval-based methods are constrained by context windows and struggle to synthesize information across long documents; \memo{} instead captures these connections via reflections during training and retrieves them through its multi-turn protocol at inference. The same trend holds on MuSiQue, where \memo{} achieves $48.30\%$ and $58.70\%$, respectively, outperforming baselines that struggle with multi-hop reasoning across independently retrieved passages. On BrowseComp-Plus, \memo{} leads with Gemini-3-Flash ($66.67\%$) and remains competitive with Qwen2.5-32B-Instruct ($54.22\%$, narrowly trailing HippoRAG2's $56.11\%$). This gap reflects BrowseComp-Plus's nature: its answers are absent from \main{}'s parametric knowledge (\cref{app:EXPR-data-contamination}), making direct access to evidence documents especially valuable and favoring retrieval methods that pass raw documents to \main{}.

\begin{table}[!ht]
    \vspace{-2mm}
    \centering
    \caption{
        Accuracy (\%) on BrowseComp-Plus, NarrativeQA, and MuSiQue under two \main{}s: Qwen2.5-32B-Instruct (Qwen2.5-32B-I) and Gemini-3-Flash (Gemini-3-F). Bold values indicate the best result in each column, excluding Perfect Retrieval. \memo{} uses Qwen2.5-14B-Instruct as \mem{}, and results are reported at the best training epoch. $^\star$Perfect Retrieval represents an empirical upper bound.
    }
    \label{tab:main-results}
    \small
    \setlength{\tabcolsep}{6pt}
    \renewcommand{\arraystretch}{1.25}
    \resizebox{\columnwidth}{!}{%
    \begin{tabular}{|l|cc|cc|cc|}
        \hline
        & \multicolumn{2}{c|}{\textbf{BrowseComp-Plus}} 
        & \multicolumn{2}{c|}{\textbf{NarrativeQA}} 
        & \multicolumn{2}{c|}{\textbf{MuSiQue}} \\
        \cline{2-3} \cline{4-5} \cline{6-7}
        \textbf{Method} 
            & \textbf{Qwen2.5-32B-I} & \textbf{Gemini-3-F} 
            & \textbf{Qwen2.5-32B-I} & \textbf{Gemini-3-F} 
            & \textbf{Qwen2.5-32B-I} & \textbf{Gemini-3-F} \\
        \hline
        Perfect Retrieval{$^\star$}
            & $79.67 \pm 1.45$ & $88.33$
            & $51.42 \pm 0.52$ & $60.41$
            & $62.83 \pm 0.90$ & $73.00$ \\
        \hline
        BM25
            & $1.11 \pm 0.69$ & $27.00$
            & $10.24 \pm 0.34$ & $14.33$
            & $20.00 \pm 0.30$ & $23.20$ \\
        NV-Embed-V2
            & $50.67 \pm 0.33$ & $57.00$
            & $20.59 \pm 0.86$ & $26.62$
            & $37.47 \pm 0.15$ & $46.60$ \\
        HippoRAG2\tablefootnote{These results differ from the original paper~\citep{gutierrez2025rag}, which uses Llama3.3-70B-Instruct instead of Qwen2.5-32B-Instruct.}
            & $\mathbf{56.11 \pm 0.51}$ & $66.33$
            & $21.39 \pm 0.20$ & $23.21$
            & $42.17 \pm 0.12$ & $57.00$ \\
        Cartridges\tablefootnote{Cartridges requires white-box access to \main{} as well; its results for Gemini-3-Flash are therefore omitted.}
            & $0.00 \pm 0.00$ & -
            & $3.75 \pm 0.11$ & -
            & $8.57 \pm 0.40$ & - \\
        \hline
        \textbf{\memo{}}
            & $54.22 \pm 0.84$ & $\mathbf{66.67}$ 
            & $\mathbf{26.85 \pm 0.39}$ & $\mathbf{53.58}$ 
            & $\mathbf{48.30 \pm 1.25}$ & $\mathbf{60.20}$ \\
        \hline
    \end{tabular}
    }
\end{table}

\para{\memo{} supports plug-and-play integration.} 
Across the three benchmarks, \memo{} consistently achieves higher performance when paired with a more capable \main{} (Gemini-3-Flash): switching from Qwen2.5-32B-Instruct to Gemini-3-Flash yields gains of 12.45\%, 26.73\%, 11.90\% on BrowseComp-Plus, NarrativeQA and MuSiQue, respectively. This demonstrates that \memo{} can be trained once with a weaker \gen{}, and seamlessly paired with \emph{any} LLM at inference, including proprietary models such as Gemini-3-Flash. This \emph{plug-and-play} capability allows \memo{} to directly leverage state-of-the-art models without any additional training or overhead.

\subsection{Ablation on the amount of noise for the dataset}
\label{subsec:noise_ablation}
\begin{table*}[!ht]
\centering
\caption{Accuracy (\%) on BrowseComp-Plus and MuSiQue with Qwen2.5-32B-Instruct as \main{}. \memo{} results are based on Qwen2.5-14B-Instruct and reported at the best training epoch. $N = N_\text{evidence}^\text{dataset}$ denotes the number of ground-truth evidence documents in the corpus; column headers indicate the number of additional negative (distractor) documents added, as a multiple of $N$. $\Delta$ denotes accuracy difference (\%) compared to $0N$.}
\label{tab:ablation-noise-robustness}
\small
\setlength{\tabcolsep}{10pt}
\renewcommand{\arraystretch}{1.25}
\resizebox{\textwidth}{!}{%
\begin{tabular}{|l|l|c|cc|}
    \hline
    \textbf{Method} & \textbf{Dataset} & \textbf{$0\times N$} & \multicolumn{2}{c|}{\textbf{$1\times N$}} \\
    & & Acc.\ (\%) & Acc.\ (\%) & $\Delta$ \\
    \hline
    \multirow{2}{*}{NV-Embed-V2} & BrowseComp-Plus & $56.89 \pm 0.51$ & $50.67 \pm 0.33$ & ${\color{red}\downarrow 6.22}$ \\
                                 & MuSiQue         & $42.30 \pm 0.53$ & $37.47 \pm 0.15$ & ${\color{red}\downarrow 4.83}$ \\
    \hline
    \multirow{2}{*}{HippoRAG2}   & BrowseComp-Plus & $62.33 \pm 1.15$ & $56.11 \pm 0.51$ & ${\color{red}\downarrow 6.22}$ \\
                                 & MuSiQue         & $47.33 \pm 0.74$ & $42.17 \pm 0.12$ & ${\color{red}\downarrow 5.16}$ \\
    \hline
    \multirow{2}{*}{\textbf{\memo{}}} & BrowseComp-Plus & $53.67 \pm 1.15$ & $54.22 \pm 0.84$ & ${\color{ForestGreen}\uparrow 0.55}$ \\
                                  & MuSiQue         & $50.07 \pm 0.81$ & $48.30 \pm 1.25$ & ${\color{orange}\downarrow 1.77}$ \\
    \hline
\end{tabular}
}
\end{table*}

We investigate the robustness of \memo{} against two strong retrieval-based baselines, NV-Embed-V2 and HippoRAG2, under increasing levels of retrieval noise, controlled by varying the number of negative (distractor) documents added to the target corpus as a multiple of the total number of ground-truth evidence documents in each dataset ($N_\text{evidence}^\text{dataset} = 1{,}775$ for BrowseComp-Plus and $N_\text{evidence}^\text{dataset} = 2{,}648$ for MuSiQue). The datasets used throughout this paper (detailed in~\cref{app:EXPR-dataset-preparation}) correspond to a ratio of 1$\times N_\text{evidence}^\text{dataset}$; we additionally evaluate at ratio 0$\times N_\text{evidence}^\text{dataset}$ (no distractors) as an idealized noise-free reference to isolate the effect of distractors.

Results in~\cref{tab:ablation-noise-robustness} demonstrate that retrieval-based methods exhibit pronounced sensitivity to noise. Both NV-Embed-V2 and HippoRAG2 suffer drops of up to 6.22\%  on BrowseComp-Plus and up to 5.16\%  on MuSiQue when scaling from $0\times N$ to $1\times N$, confirming that these systems struggle to filter irrelevant documents under realistic corpus conditions. In contrast, \memo{} maintains stable performance across both benchmarks, with a marginal improvement of 0.55\%  on BrowseComp-Plus and a decline of only 1.77\%  on MuSiQue, both within one standard deviation, demonstrating that \memo{} is robust to increasing retrieval noise. We attribute this robustness to \memo{}'s design: despite being trained on a corpus containing negative documents, \mem{} provides more precise information to \main{}'s sub-queries than direct document retrieval. Additional analysis of performance degradation in retrieval-based methods is provided in~\cref{app:perf-degradation-with-noise}.

\subsection{Ablation on \mem{} size}
\label{subsec:ablate-mem-size}
We investigate how the size of \mem{} affects downstream task performance by comparing models of 1.5B and 14B parameters in the Qwen2.5 family. Implementation details are provided in~\cref{app:ablate-mem-size}.
Results in~\cref{tab:with-N-noise-ablation-mem-size} show a consistent positive scaling trend: larger \mem{}s yield improved performance across all benchmarks and \main{}s. However, the results also show that a stronger \main{} reasoning capability modulates this gap non-uniformly across tasks: the performance difference between \mem{} sizes widens for NarrativeQA but shrinks for BrowseComp-Plus and MuSiQue. This suggests that the interaction between \main{} reasoning capability and \mem{} size is task-dependent.

\begin{table}[!ht]
    \vspace{-4mm}
    \centering
    \caption{
        Ablation on \mem{} size within the Qwen2.5 family. Bold results indicate best performing results in the column.
    }
    \label{tab:with-N-noise-ablation-mem-size}
    \small
    \setlength{\tabcolsep}{6pt}
    \renewcommand{\arraystretch}{1.25}
    \resizebox{\columnwidth}{!}{%
    \begin{tabular}{|c|cc|cc|cc|}
        \hline
        & \multicolumn{2}{c|}{\textbf{BrowseComp-Plus}} 
        & \multicolumn{2}{c|}{\textbf{NarrativeQA}} 
        & \multicolumn{2}{c|}{\textbf{MuSiQue}} \\
        \textbf{\memory{} Model} 
        & \textbf{Qwen2.5-32B} & \textbf{Gemini-3-Flash}
        & \textbf{Qwen2.5-32B} & \textbf{Gemini-3-Flash}
        & \textbf{Qwen2.5-32B} & \textbf{Gemini-3-Flash} \\ \hline
        Qwen2.5-1.5B-Instruct  & $44.11 \pm 2.22$ & $61.00$ & $24.00 \pm 0.20$ & $47.44$ & $42.90 \pm 1.39$ & $59.70$ \\ \hline
        Qwen2.5-14B-Instruct   & $\mathbf{54.22 \pm 0.84}$ & $\mathbf{66.67}$ & $\mathbf{26.85 \pm 0.39}$ & $\mathbf{53.58}$ & $\mathbf{48.30 \pm 1.25}$ & $\mathbf{60.20}$ \\ \hline
    \end{tabular}
    }
\end{table}

\subsection{Ablation on \mem{} family}
We investigate whether the choice of \mem{} family affects performance by comparing three models of similar parameter scale ($\sim$1–2B) but distinct architectures and pretraining lineages: Qwen2.5-1.5B-Instruct~\citep{qwen2025qwen25technicalreport}, Gemma3-1B-IT~\citep{gemmateam2025gemma3technicalreport}, and LFM2.5-1.2B-Instruct~\citep{amini2025lfm2}. Implementation details are provided in~\cref{app:ablate-mem-family}.
Results in~\cref{tab:with-N-noise-ablation-mem-family} show that \memo{} performance is largely robust to the choice of \mem{} architecture, demonstrating that the framework is not sensitive to the specific pretraining lineage of \mem{} at similar parameter scale, 
and that the parametric knowledge compression induced by our training procedure generalizes across diverse model families.

\begin{table}[H]
    \vspace{-4mm}
    \centering
    \caption{
        Ablation across \mem{}s at similar parameter scales ($\sim$1–2B). Bold results indicate best performing results in the column.
    }
    \label{tab:with-N-noise-ablation-mem-family}
    \small
    \setlength{\tabcolsep}{6pt}
    \renewcommand{\arraystretch}{1.25}
    \resizebox{\columnwidth}{!}{%
    \begin{tabular}{|c|cc|cc|cc|}
        \hline
        & \multicolumn{2}{c|}{\textbf{BrowseComp-Plus}} 
        & \multicolumn{2}{c|}{\textbf{NarrativeQA}} 
        & \multicolumn{2}{c|}{\textbf{MuSiQue}} \\
        \textbf{\memory{} Model} 
        & \textbf{Qwen2.5-32B-I} & \textbf{Gemini-3-F}
        & \textbf{Qwen2.5-32B-I} & \textbf{Gemini-3-F}
        & \textbf{Qwen2.5-32B-I} & \textbf{Gemini-3-F} \\ \hline
        Qwen2.5-1.5B-Instruct  & $\mathbf{44.11 \pm 2.22}$ & $\textbf{61.00}$ & $\mathbf{24.00 \pm 0.20}$ & $47.44$ & $42.90 \pm 1.39$ & $\mathbf{59.70}$ \\ \hline
        Gemma3-1B-IT           & $41.67 \pm 2.03$ & $59.00$ & $22.30 \pm 2.47$ & $\mathbf{48.81}$ & $41.17 \pm 1.20$ & $56.20$ \\ \hline
        LFM2.5-1.2B-Instruct   & $37.33 \pm 1.86$ & $59.67$ & $21.96 \pm 1.97$ & $46.42$ & $\mathbf{45.23 \pm 2.49}$ & $58.30$ \\ \hline
    \end{tabular}
    }
    \vspace{-2mm}
\end{table}

\subsection{Continual integration via model merging}
\label{sec:experiments-merging}

We test the streaming-update scenario described in \cref{sec:training} on NarrativeQA, comparing model merging against full retraining of \mem{} on the union of both subsets when the second arrives. Of the 14 sweep configurations (see \cref{tab:merge-method-sweep}, \cref{app:ALGO-model-training-discussion}), we report TIES~\citep{yadav2023ties} at $\rho{=}0.3$ in the main paper, the top-performing one. Letting $X$ and $Y$ denote the SFT cost on each subset alone (cost scales approximately linearly with the number of QA pairs, so the union costs $X{+}Y$), cumulative compute across the two arrivals is $X{+}Y$ for merging versus $X{+}(X{+}Y)$ for full retraining.

\begin{table}[!ht]
    \vspace{-2mm}
    \centering
    \caption{
        \textbf{Model merging vs.\ full retraining on NarrativeQA.}
        \mem{} = Qwen2.5-14B-Instruct. Merge-TIES ($\rho{=}0.3$) is the best of 14 configurations swept (\cref{tab:merge-method-sweep},
        \cref{app:ALGO-model-training-discussion}). Cumulative compute is reported in 8$\times$H100 GPU-hours for $K{=}2$ subsets of $\sim$640k reflection QA pairs each. $\Delta$ denotes accuracy difference (\%) relative to full retraining.
    }
    \label{tab:merging-results}
    \small
    \setlength{\tabcolsep}{6pt}
    \renewcommand{\arraystretch}{1.25}
    \begin{tabular}{|l|c|cc|cc|}
        \hline
        \textbf{Method} & \textbf{Cumulative compute}
                        & \multicolumn{2}{c|}{\textbf{Qwen2.5-32B-I}}
                        & \multicolumn{2}{c|}{\textbf{Gemini-3-F}} \\
        & (8$\times$H100 GPU-h) & Acc.\ (\%) & $\Delta$ & Acc.\ (\%) & $\Delta$ \\
        \hline
        Full retrain ($X{+}(X{+}Y)$)
            & $\approx 72$h
            & $\mathbf{26.85 \pm 0.39}$ & ---
            & $\mathbf{53.58}$          & --- \\
        Merge-TIES ($\rho{=}0.3$, $X{+}Y$)
            & $\approx 48$h
            & $15.81 \pm 0.39$ & ${\color{red}\downarrow 11.04}$
            & $34.47$          & ${\color{red}\downarrow 19.11}$ \\
        \hline
    \end{tabular}
\end{table}

\para{Merging cuts compute by $\mathbf{33\%}$ at $K{=}2$, with widening returns at scale.}
As reported in \cref{tab:merging-results}, the full-retrain baseline incurs $X{+}(X{+}Y) \approx 72$ GPU-hours of cumulative compute, while merging accumulates only $X{+}Y \approx 48$ GPU-hours --- a $33\%$ reduction (\cref{fig:cost_vs_accuracy}). The gap widens with $K$: under the same per-corpus cost, merging scales as $\Theta(K)$ while full retraining scales as $\Theta(K^2)$, yielding a $5.5{\times}$ saving at $K{=}10$ ($240$ vs.\ $1{,}320$ GPU-hours).

\para{Merging trades a measurable accuracy gap for the compute saving, but still beats retrieval.}
Merge-TIES ($\rho{=}0.3$) trails the full-retrain \mem{} by $11.0$\%  under Qwen2.5-32B-Instruct and $19.1$\%  under Gemini-3-Flash (\cref{tab:merging-results}); across the full 14-configuration sweep, accuracy ranges from $7.85\%$ (SLERP, worst) to $15.81\%$ (TIES, best), shown in \cref{fig:cost_vs_accuracy}. Despite this gap, the merged \mem{} still outperforms every retrieval baseline (BM25, NV-Embed-V2, HippoRAG2, Cartridges; see \cref{tab:main-results}) on NarrativeQA, indicating that even an aggressively-cheaper merging procedure preserves most of \memo{}'s qualitative advantage over retrieval-based approaches. TIES and DARE-Linear at $\rho{=}0.3$ dominate the sweep, suggesting that aggressive sparsification combined with sign-conflict resolution is the most reliable merging recipe in this regime.

%% file: latex/conclusion.tex

We introduced \memo{}, a modular framework for integrating updated or domain-specific knowledge into LLMs via a \mem{} trained on a synthesized reflection QA dataset. \memo{} addresses key limitations of existing methods: it bypasses context constraints and limited cross-document reasoning in retrieval-based approaches, avoids costly and brittle parametric updates (including catastrophic forgetting), and removes representation coupling in latent memory methods. Its core components are a data synthesis pipeline capturing explicit facts and implicit relationships, and a multi-turn inference protocol that decomposes complex queries into targeted sub-queries for desired information retrieval from the memory model.
While \memo{} demonstrates strong performance, it has limitations regarding training cost, evaluation scope, and the capacity of \mem{} to scale with corpus size (see \cref{app:limitations}).
Empirically, \memo{} outperforms strong baselines across diverse benchmarks. It also provides a scalable pathway for knowledge integration, supporting efficient updates and plug-and-play deployment with both open and proprietary closed-source LLMs.
Future work includes more efficient memory construction, extensions to dynamic corpora, and tighter coordination between the \main{} and \mem{}. We view \textbf{\memo{}} (\textbf{Me}mory as a \textbf{Mo}del) as a promising foundation for more flexible, updatable, and knowledge-aware AI systems.

%% file: latex/appendix.tex

\crefalias{section}{appendix}
\crefalias{subsection}{appendix}   
\crefalias{subsubsection}{appendix}

\section{Impact statement}
\label{app:broader-impacts}
\memo{} advances the ability of LLMs to internalize knowledge over large, domain-specific corpora without requiring access to model weights, lowering the barrier for deploying capable AI systems in knowledge-intensive domains such as law, medicine, and scientific research. By enabling plug-and-play integration with any LLM, including proprietary models, \memo{} democratizes access to powerful knowledge integration capabilities that would otherwise require significant computational resources or white-box model access. At the same time, this accessibility introduces dual-use concerns, as the same capability that enables beneficial applications could be used to internalize misinformation, proprietary data without authorization, or harmful content at scale. Additionally, as \memo{} reduces reliance on explicit retrieval, it may obscure the provenance of retrieved information, making it harder to attribute the sources underlying a model's responses. We encourage future work to investigate attribution mechanisms and access controls for memory-based systems, and urge practitioners to carefully consider the nature of the documents used to train \mem{}.

\section{Limitations}
\label{app:limitations}
\memo{} incurs an upfront training cost for each new corpus, and performance may vary across domains, document types, or LLM families beyond those covered in our experiments. Furthermore, the performance of \memo{} is inherently bounded by the representational capacity of \mem{} to internalize the target corpus. Although our experiments do not reveal clear signs that \mem{} has reached its capacity limit, we hypothesize that sufficiently large or information-dense corpora will exceed what a fixed-size \mem{} can correctly compress and represent.

\section{Future work}
We outline several directions for future work. The data generation pipeline is computationally expensive, with Step~5 in \cref{alg:datagen-pipeline} scaling quadratically at $O(k \cdot C^2 \cdot Q^2)$, and reducing this cost remains an open problem. A systematic evaluation of chunking strategies and their associated tradeoffs~(\cref{app:EXPR-dataset-preparation}) is likewise an open direction. On the training side, scaling \mem{} with corpus size and developing more effective model merging strategies for reducing per-corpus training costs~(\cref{sec:experiments-merging}) are promising directions. Other post-training methods such as Reinforcement Learning~\citep{sutton1998reinforcement} have also shown to be effective in improving model task performance~\citep{lambert2024tulu}, and applying such methods to \mem{} training warrants future investigation. 

LoRA configurations better suited to specific architectures, including per-architecture tuning of rank and learning rate, also warrant further investigation~(\cref{app:EXPR-sft-vs-lora}). Finally, a more systematic study of the interaction between \main{} reasoning capability and \mem{} model size~(\cref{subsec:ablate-mem-size}), as well as the optimal interaction budget at each stage and \main{} selection~(\cref{app:eval-setup-ablation}), are other promising future directions.

\section{Preparation of datasets}
\label{app:EXPR-dataset-preparation}

\para{Corpus construction.}
Extending from our description in \cref{sec:experiments}, we distinguish between two types of documents\footnote{Note that for BrowseComp-Plus, the gold documents are a subset of the evidence documents.}: evidence documents, which contain information relevant to answering a given question, and negative documents, which are irrelevant and serve as noise. For BrowseComp-Plus, we used $1{,}775$ unique evidence documents and $1{,}766$ unique negative documents (after removal of non-English documents), yielding $3{,}541$ documents in total. For MuSiQue, we used $2{,}648$ documents for each of the evidence and negative documents, yielding $5{,}296$ documents in total. NarrativeQA does not have negative documents.

\para{Chunking strategy.}
As shown in~\cref{tab:token-distribution}, NarrativeQA full documents span the $32{,}769$--$131{,}072$ token range with a median length of $65{,}925$ tokens, reflecting the long-form nature of the source novels. Processing such documents without chunking risks reduced coverage of extractable QA pairs in Step~1 of~\cref{alg:datagen-pipeline}, as attention quality is known to deteriorate over longer contexts~\citep{hsieh2024ruler}. We therefore chunk NarrativeQA documents using a fixed sliding window of $6{,}400$ words with a $640$-word overlap ($10\%$ overlap ratio), yielding $75$ chunks concentrated in the $4{,}097$--$16{,}384$ token range and accounting for $96\%$ of all chunks, with a median group size of $7$ per document as shown in~\cref{tab:group-distribution}. Unlike NarrativeQA, MuSiQue documents are compact with $99.70\%$ falling below $512$ tokens, and each MuSiQue document is treated as a single chunk.

\begin{table}[!ht]
    \vspace{-4mm}
    \centering
    \caption{Token length distribution across corpora at the \textit{chunk level}, where $n$ represents the total number of individual chunks processed by~\cref{alg:datagen-pipeline}. Each entry reflects the token count of a single text chunk. Statistics for NarrativeQA are reported before and after chunking.}
    \label{tab:token-distribution}
    \small
    \setlength{\tabcolsep}{6pt}
    \renewcommand{\arraystretch}{1.25}
    \resizebox{\columnwidth}{!}{%
    \begin{tabular}{|l|r|r|r|r|}
        \hline
        \textbf{Token Range} 
        & \thead{\textbf{BrowseComp-Plus} \\ \textbf{($n=3{,}541$)}} 
        & \thead{\textbf{NarrativeQA} \\ \textbf{Full Docs} \\ \textbf{($n=10$)}} 
        & \thead{\textbf{NarrativeQA} \\ \textbf{Chunks} \\ \textbf{($n=75$)}} 
        & \thead{\textbf{MuSiQue} \\ \textbf{($n=5{,}296$)}} \\ \hline
        $0$--$512$               & $606$ ($17.11\%$)  & $0$ ($0.00\%$)    & $0$ ($0.00\%$)     & $5{,}280$ ($99.70\%$) \\ \hline
        $513$--$1{,}024$         & $591$ ($16.69\%$)  & $0$ ($0.00\%$)    & $0$ ($0.00\%$)     & $16$ ($0.30\%$)       \\ \hline
        $1{,}025$--$2{,}048$     & $746$ ($21.07\%$)  & $0$ ($0.00\%$)    & $1$ ($1.33\%$)     & $0$ ($0.00\%$)        \\ \hline
        $2{,}049$--$4{,}096$     & $598$ ($16.89\%$)  & $0$ ($0.00\%$)    & $2$ ($2.67\%$)     & $0$ ($0.00\%$)        \\ \hline
        $4{,}097$--$8{,}192$     & $428$ ($12.09\%$)  & $0$ ($0.00\%$)    & $36$ ($48.00\%$)   & $0$ ($0.00\%$)        \\ \hline
        $8{,}193$--$16{,}384$    & $323$ ($9.12\%$)   & $0$ ($0.00\%$)    & $36$ ($48.00\%$)   & $0$ ($0.00\%$)        \\ \hline
        $16{,}385$--$32{,}768$   & $145$ ($4.09\%$)   & $0$ ($0.00\%$)    & $0$ ($0.00\%$)     & $0$ ($0.00\%$)        \\ \hline
        $32{,}769$--$65{,}536$   & $56$ ($1.58\%$)    & $5$ ($50.00\%$)   & $0$ ($0.00\%$)     & $0$ ($0.00\%$)        \\ \hline
        $65{,}537$--$131{,}072$  & $20$ ($0.56\%$)    & $5$ ($50.00\%$)   & $0$ ($0.00\%$)     & $0$ ($0.00\%$)        \\ \hline
        $>131{,}072$             & $28$ ($0.79\%$)    & $0$ ($0.00\%$)    & $0$ ($0.00\%$)     & $0$ ($0.00\%$)        \\ \hline
        \textbf{Min tokens}      & $14$              & $32{,}804$        & $1{,}943$          & $23$                  \\ \hline
        \textbf{Median tokens}   & $1{,}756$         & $65{,}925$        & $8{,}158$          & $105$                 \\ \hline
        \textbf{Mean tokens}     & $7{,}192$         & $66{,}324$        & $8{,}713$          & $123$                 \\ \hline
        \textbf{$p_{95}$ tokens} & $20{,}330$        & $119{,}267$       & $11{,}266$         & $270$                 \\ \hline
        \textbf{Max tokens}      & $1{,}235{,}897$   & $119{,}267$       & $12{,}104$         & $828$                 \\ \hline
    \end{tabular}%
    }
\end{table}

\begin{table}[!ht]
    \vspace{-4mm}
    \centering
    \caption{Distribution of document group sizes across datasets, where group size denotes the number of chunks associated with a single question or document. For BrowseComp-Plus and MuSiQue, each question is associated with a subset of chunks drawn from the corpus, and group size represents the number of chunks per \emph{question}. For NarrativeQA, each subset of chunks is derived from the original document used for multiple questions, and group size represents the number of chunks per \emph{document}.}
    \label{tab:group-distribution}
    \small
    \setlength{\tabcolsep}{6pt}
    \renewcommand{\arraystretch}{1.25}
    \resizebox{\columnwidth}{!}{%
    \begin{tabular}{|l|r|r|r|}
        \hline
        \thead{\textbf{Document Group} \\ \textbf{Size Range}} 
        & \thead{\textbf{BrowseComp-Plus} \\ \textbf{($n_{\text{group}} = 300$)}} 
        & \thead{\textbf{NarrativeQA Chunks} \\ \textbf{($n_{\text{group}} = 10$)}} 
        & \thead{\textbf{MuSiQue} \\ \textbf{($n_{\text{group}} = 1{,}000$)}} \\ \hline
        0--2   & $2$ ($0.67\%$)   & $0$ ($0.00\%$)  & $0$ ($0.00\%$)    \\ \hline
        3--4   & $14$ ($4.67\%$)  & $3$ ($30.00\%$) & $518$ ($51.80\%$) \\ \hline
        5--8   & $78$ ($26.00\%$) & $4$ ($40.00\%$) & $482$ ($48.20\%$) \\ \hline
        9--16  & $159$ ($53.00\%$)& $3$ ($30.00\%$) & $0$ ($0.00\%$)    \\ \hline
        $>$16  & $47$ ($15.67\%$) & $0$ ($0.00\%$)  & $0$ ($0.00\%$)    \\ \hline
        \textbf{Min group size}      & $2$    & $3$   & $4$   \\ \hline
        \textbf{Median group size}   & $12$   & $7$   & $4$   \\ \hline
        \textbf{Mean group size}     & $11.8$ & $7.5$ & $5.3$ \\ \hline
        \textbf{$p_{95}$ group size} & $20$   & $16$  & $8$   \\ \hline
        \textbf{Max group size}      & $23$   & $16$  & $8$   \\ \hline
    \end{tabular}%
    }
\end{table}

BrowseComp-Plus documents are also treated as a single chunk. The time complexity of Step~5 in \cref{alg:datagen-pipeline} is $O(k \cdot C^2 \cdot Q^2)$, where $k = n_{\text{group}}$ is the number of groups, $C = |G_i|$ is the number of participating chunks per group, and $Q = \bar{Q}_i$ is the average number of QA pairs extracted per chunk. Since chunking increases $C$, pipeline costs at Step~5 scale quadratically as the number of chunks per group increases. Given that only $2.93\%$ of BrowseComp-Plus documents exceed $32{,}768$ tokens, the majority of documents fit within a single chunk, making the cost of chunking difficult to justify. We therefore opted against chunking in favor of lower pipeline cost, and leave a systematic evaluation of chunking strategies and related tradeoffs to future work.

\para{Subset selection of negative documents.} 
We include only a subset of negative documents for BrowseComp-Plus and MuSiQue due to computational constraints arising from the quadratic scaling of Step~5. As reported in~\cref{tab:group-distribution}, BrowseComp-Plus currently has a mean group size of $11.8$ and a maximum of $23$, while MuSiQue has a mean group size of $5.3$ and a maximum of $8$. Incorporating all available negative documents, which average $78$ per question (up to $197$) for BrowseComp-Plus and $17$ per question (up to $18$) for MuSiQue, would cause the group size to increase substantially. Given the quadratic dependence on $C$ in Step~5, this would result in a prohibitive increase in pipeline cost for BrowseComp-Plus ($k = 300$) and MuSiQue ($k = 1{,}000$). Hence, we opted to only include up to $N_\text{evidence}^\text{dataset}$ number of negative documents for each question in the corpus.

\section{Discussion on steps in data generation pipeline}
\label{app:datagen-pipeline-Ablation}
\subsection{Ablation of data synthesis steps}
\label{app:datagen-current-step-ablation}
We experiment with the data generation pipeline to show the importance of each step. We perform a leave-one-out (LOO) ablation for each step of data synthesis and train the model on the synthesized QA pairs generated. Results are reported in~\cref{tab:with-N-noise-datagen-ablation} on the NarrativeQA and MuSiQue datasets using Qwen2.5-32B-Instruct as the \main{} and Qwen2.5-1.5B-Instruct as the \mem{}.

\begin{table}[!ht]
    \vspace{-4mm}
    \centering
    \caption{LOO ablation accuracy at best performing Qwen2.5-1.5B-Instruct epoch across datasets. Data ratio indicates the number of QA pairs retained relative to the baseline. For each step removed, the Qwen2.5-1.5B-Instruct was retrained, and we report the mean $\pm$ std. dev. over 3 runs at the same training epoch as the baseline.}
    \label{tab:with-N-noise-datagen-ablation}
    \small
    \setlength{\tabcolsep}{6pt}
    \renewcommand{\arraystretch}{1.25}
    \resizebox{\columnwidth}{!}{%
    \begin{tabular}{|c|cc|cc|}
        \hline
        & \multicolumn{2}{c|}{\textbf{NarrativeQA}} 
        & \multicolumn{2}{c|}{\textbf{MuSiQue}} \\
        \textbf{Ablation} 
        & \textbf{Data Ratio} & \textbf{Accuracy (\%)}
        & \textbf{Data Ratio} & \textbf{Accuracy (\%)} \\ \hline
        Baseline (all steps) & $1.000\times$ & $24.00 \pm 0.20$ & $1.000\times$ & $42.90 \pm 1.39$ \\ \hline
        Step~1a removed             & $0.434\times$ & $20.48 \pm 0.90$ & $0.381\times$ & $30.00 \pm 0.17$ \\ \hline
        Step~1b removed             & $0.598\times$ & $22.98 \pm 1.04$ & $0.651\times$ & $37.33 \pm 0.25$ \\ \hline
        Step~2 removed              & $0.739\times$ & $24.69 \pm 1.10$ & $0.621\times$ & $37.10 \pm 1.76$ \\ \hline
        Step~3 removed              & $2.078\times$ & $28.90 \pm 0.86$ & $1.128\times$ & $41.70 \pm 0.78$ \\ \hline
        Step~4 removed              & $0.378\times$ & $23.21 \pm 1.56$ & $0.501\times$ & $39.10 \pm 0.02$ \\ \hline
        Step~5 removed              & $0.002\times$ & $6.37  \pm 0.39$ & $0.195\times$ & $24.17 \pm 0.25$ \\ \hline
    \end{tabular}%
    }
\end{table}

Step~5 (Cross-document synthesis) is the most critical component of the pipeline. Its removal causes accuracy to collapse to $6.37\%$ and $24.17\%$ on NarrativeQA and MuSiQue respectively, against baseline scores of $24.00\%$ and $42.90\%$, accompanied by a near-total loss of training data ($0.002\times$ and $0.195\times$ retention). As described in \cref{sec:datagen}, Step~5 enables cross-document synthesis where $\cM_{\text{gen}}$ constructs $\cQ_{\text{cross}}$ pairs spanning inter-document connections and cross-chunk connections within a single long document, making it the dominant source of training pairs in $\cQ_{\text{final}}$ and directly targeting the multi-source synthesis objective central to both benchmarks.

An interesting anomaly arises with Step~2 and Step~3, where their removal does not consistently hurt performance and improves accuracy on NarrativeQA. Step~2 merges related QA pairs from a single document chunk into multi-fact questions by identifying commonalities such as shared entities, overlapping time periods, and sequential events. For MuSiQue, these commonalities reflect genuine knowledge relationships that directly resemble the multi-hop factual reasoning the benchmark evaluates, such that removing Step~2 eliminates a large fraction of useful training pairs, leading to a drop in accuracy from $42.90\%$ to $37.10\%$. For NarrativeQA, however, the same consolidation patterns operate on superficial narrative co-occurrences rather than meaningful knowledge relationships. The predominant commonality categories are event or scene groupings that NarrativeQA does not evaluate, and entity co-occurrence patterns that are trivially satisfied given the pervasive presence of central characters across scenes. Removing Step~2 eliminates these low-quality pairs, leading to the marginal accuracy improvement from $24.00\%$ to $24.69\%$.

Removing Step~3 retains more data than the baseline ($2.078\times$ and $1.100\times$ for NarrativeQA and MuSiQue respectively), yet the effect on performance diverges. For MuSiQue, performance drops from $42.90\%$ to $41.78\%$, whereas for NarrativeQA, performance improves from $24.00\%$ to $28.90\%$. Step~3 applies a self-containment filter that rewrites or discards pairs whose questions cannot be understood without access to the source chunk. 
For MuSiQue, violations are predominantly localized and shallow, making them amenable to filtering; the proposed filter effectively identifies and removes defective pairs.
For NarrativeQA, long-form narrative text frequently contains pronouns and temporal references that span many paragraphs, which are structural features of the domain rather than fixable defects. This causes the rewriting loop to introduce substitute unrelated content and corrupt the pairs produced by earlier steps. Removing Step~3 for NarrativeQA therefore avoids this domain-induced corruption and retains the original pairs intact, explaining both the data retention ratio increase and the accuracy improvement. This suggests that Step~3 is most beneficial when applied to domains where self-containment violations are well-defined and resolvable.

The remaining steps follow a consistent trend: removing Step~1a, Step~1b, or Step~4 reduces both data volume and accuracy across both datasets, confirming that each step contributes a distinct and meaningful role to the final training corpus quality.

\subsection{Additional steps considered but excluded}
Three additional steps were considered but ultimately excluded from the pipeline. These include paraphrasing~\citep{ovadia2024fine}, increasing the number of sampling trials at Step~1 of~\cref{alg:datagen-pipeline}, and a targeted fill whereby $\cM_{\text{gen}}$ reviews the generated QA pairs and rewrites them to incorporate additional missed information. Paraphrasing was excluded as the scale of generated pairs already provides sufficient coverage (\(\approx\)600k--1.6M across the three datasets, see Table~\ref{tab:batch_sizes}), and the potential gains were outweighed by the additional computational overhead. Increasing sampling trials proved unreliable, as additional trials did not consistently extract facts that the initial pass had failed to extract. The targeted fill similarly offered limited gains, where appending the existing QA pairs as context to prompt a revision only lengthens the context when the model had already failed to extract a fact from the original chunk, likely exacerbating attention further degradation over long inputs~\citep{hsieh2024ruler} and making retrieval of relevant information less reliable at inference time.

\section{\memory{} model hyperparameter settings}
\label{app:ALGO-train-settings}
Training was conducted on H100 and H200 GPUs using the hyperparameter settings reported in~\cref{tab:training_config}. The effective batch size for each dataset is summarized in~\cref{tab:batch_sizes}.

\begin{table}[!ht]
    \vspace{-4mm}
    \centering
    \caption{\mem{} SFT Training Configuration}
    \label{tab:training_config}
    \small
    \setlength{\tabcolsep}{6pt}
    \renewcommand{\arraystretch}{1.25}
    \begin{tabular}{|l|c|}
        \hline
        \textbf{Parameter} & \textbf{Value} \\
        \hline
        Optimizer & Fused AdamW \\
        \hline
        Gradient checkpointing & True \\
        \hline
        Learning rate (LR) & $2 \times 10^{-5}$ \\
        \hline
        Num of Training epochs & 3 \\
        \hline
        LR scheduler type & Constant with warmup \\
        \hline
        Warmup ratio & 0.05 \\
        \hline
        Weight decay & 0.01 \\
        \hline
        Max gradient norm & 1.0 \\
        \hline
        Max sequence length & 8096 \\
        \hline
        Precision & BF16 \\
        \hline
        Attention implementation & Flash Attention 2 \\
        \hline
    \end{tabular}
\end{table}

\begin{table}[!ht]
    \vspace{-4mm}
    \centering
    \caption{Effective batch sizes and number of QA pairs used. NarrativeQA.1 and NarrativeQA.2 are independent subsets partitioned from the original that were used for model merging.}
    \label{tab:batch_sizes}
    \small
    \setlength{\tabcolsep}{6pt}
    \renewcommand{\arraystretch}{1.25}
    \begin{tabular}{|l|c|c|c|}
    \hline
    \textbf{Dataset} & \textbf{Target Num of Questions} & \textbf{Num of QA Pairs} & \textbf{Effective Batch Size} \\
    \hline
    BrowseComp-Plus & $300$ & $1{,}639{,}995$ & $512$ \\
    \hline
    NarrativeQA & $293$ & $1{,}276{,}676$ & $512$ \\
    \hline
    NarrativeQA.1 & $146$ & $635{,}009$ & $256$ \\
    \hline
    NarrativeQA.2 & $147$ & $641{,}667$ & $256$ \\
    \hline
    MuSiQue     & $1{,}000$ & $664{,}762$ & $256$ \\
    \hline
    \end{tabular}
\end{table}

\section{Compute resources}
\label{app:EXPR-compute-resources}

All experiments were conducted using NVIDIA H200 GPUs. We report computational cost in GPU-hours.
\paragraph{Data generation.} Generating the full reflection dataset for BrowseComp-Plus, NarrativeQA, and MuSiQue took approximately 240, 200, and 150 GPU-hours respectively.
\paragraph{Training.} \mem{} (Qwen2.5-14B-Instruct) training for a single run BrowseComp-Plus, NarrativeQA, and MuSiQue took approximately 180, 150, 90 GPU-hours.

\section{Model training discussion} \label{app:ALGO-model-training-discussion}
We considered three training paradigms: CPT, SFT, and LoRA-based SFT. CPT was excluded as it risks degrading instruction-following capability~\citep{wu2024continual}, which is critical for downstream QA evaluation. Full SFT was selected as it directly optimizes for the target task while preserving alignment~\citep{ouyang2022training}. LoRA-based SFT serves as a parameter-efficient alternative and we include a comparison to these training methods in~\cref{app:EXPR-sft-vs-lora}.

Model merging targets the practical streaming setting in which new corpora arrive over time and \mem{} must continually integrate them. Retraining \mem{} from scratch on the union of all observed corpora is the natural baseline but quickly becomes prohibitive at scale, since its cost grows with the cumulative corpus size. Model merging instead trains a separate \mem{} on each new corpus and combines it with the existing model in parameter space, so the cost of each update scales only with the size of the new corpus rather than the entire history. This decoupling comes at a measurable accuracy cost relative to full retraining, which we quantify in \cref{fig:cost_vs_accuracy}. We assume the corpora to be merged are pairwise disjoint.

\subsection{Model merging}
\label{app:model-merging}

\para{Merging methods.}
We consider the following methods, all of which produce $\varphi_{\text{merged}}$ without ever training on $\mathcal{D}_1 \cup \dots \cup \mathcal{D}_K$:
\begin{itemize}
    \setlength\itemsep{0.0em}
  \item \textbf{Linear merging}~\citep{wortsman2022model} computes a weighted sum of task vectors: $\varphi_{\text{merged}} = \varphi_0 + \sum_{i=1}^{K} \lambda_i \tau_i$, where $\lambda_i > 0$ are merging coefficients.
  \item \textbf{SLERP}~\citep{shoemake1985animating} interpolates between two task vectors along the unit sphere, preserving their magnitudes: $\varphi_{\text{merged}} = \varphi_0 + \mathrm{SLERP}(\tau_1, \tau_2;\, t)$, with $t \in [0, 1]$ controlling the interpolation factor.
  \item \textbf{Task arithmetic}~\citep{ilharco2023editing} adds task vectors directly without further processing, recovering linear merging as a special case with uniform $\lambda_i$.
  \item \textbf{TIES}~\citep{yadav2023ties} resolves interference among task vectors before summation by (i) \emph{trimming} each $\tau_i$ to its top-$\rho$ fraction of largest-magnitude entries, (ii) \emph{electing} a sign at each coordinate by magnitude-weighted majority vote, and (iii) \emph{disjoint-merging} only the entries that agree with the elected sign.
  \item \textbf{DARE}~\citep{yu2024language} sparsifies each task vector by randomly dropping a fraction $1 - \rho$ of its entries and rescaling the survivors by $1 / \rho$ to preserve expected magnitude, before linear merging.
  \item \textbf{DARE-TIES}~\citep{yu2024language} combines DARE-style stochastic sparsification with TIES sign-conflict resolution, retaining the diversity of random dropout while filtering out conflicting updates.
\end{itemize}

\para{Avoiding catastrophic forgetting.}
Because no individual \mem{} $\mathcal{M}_{\varphi_i}$ is ever fine-tuned on another corpus' data, model merging cannot induce the kind of distributional interference that drives catastrophic forgetting in sequential fine-tuning~\citep{luo2025empiricalstudycatastrophicforgetting}. Knowledge from each corpus is preserved within its own task vector $\tau_i$, and conflicts between task vectors are addressed at merge time via the methods above rather than during gradient updates.

\para{Scalability.}
When a new corpus $\mathcal{D}_{K+1}$ arrives, we train auxiliary model $\mathcal{M}_{\varphi_{K+1}}$ on its reflection QA dataset, derive $\tau_{K+1}$, and re-merge in $\mathcal{O}(1)$ additional cost relative to the full collection. This enables modular, plug-and-play integration over a continuous stream of disjoint knowledge sources, unlike retraining from scratch on $\bigcup_i \mathcal{D}_i$, which scales linearly with the cumulative corpus size.

\para{Inference.}
The merged \mem{} is queried identically to a single-corpus \mem{} via the structured multi-turn protocol described in \cref{sec:inference}. Because merging operates entirely in parameter space and produces a model with the same architecture and interface as $\mathcal{M}_{\varphi_0}$, it inherits the plug-and-play property of \memo{} without requiring changes to the \main{} or the inference protocol. Importantly, the \main{} queries a \emph{single} merged \mem{} at inference rather than dispatching across $K$ separate per-corpus \mem{}s, keeping the multi-turn retrieval pipeline unchanged regardless of how many corpora have been integrated.

\para{Procedure.}
For our experiments we partition NarrativeQA into two pairwise-disjoint subsets, NarrativeQA.1 and NarrativeQA.2, of $\sim$640k reflection QA pairs each. Each subset is used to fine-tune an independent \mem{} from the same Qwen2.5-14B-Instruct base via SFT for $3$ epochs, producing $\mathcal{M}_{\varphi_1}$ and $\mathcal{M}_{\varphi_2}$ at SFT costs of $X$ and $Y$ GPU-hours, respectively (each is $\approx 24$ GPU-hours on 8$\times$H100; full-retrain on the union NarrativeQA.1 $\cup$ NarrativeQA.2 costs $X{+}Y \approx 48$ GPU-hours by linear scaling). We evaluate every saved checkpoint of each run on the held-out NarrativeQA evaluation set and select the best-performing checkpoint per subset; the corresponding task vectors $\tau_1$ and $\tau_2$ are the inputs to the merging step. We then sweep all six merging methods listed above (Linear, Task arithmetic, SLERP, TIES, DARE, DARE-TIES) at three sparsification densities $\rho \in \{0.3, 0.5, 0.7\}$ (or three interpolation factors $t \in \{0.3, 0.5, 0.7\}$ for SLERP), giving $14$ merged-\mem{} configurations in total. Each configuration is evaluated on NarrativeQA with Qwen2.5-32B-Instruct as \main{} (mean $\pm$ std over 3 runs). The configuration that we report in the \cref{sec:experiments} as \emph{Merge-TIES} is the best of the sweep (TIES with $\rho{=}0.3$).

\begin{figure}[!ht]
    \vspace{-3mm}
    \centering
    \includegraphics[width=\linewidth]{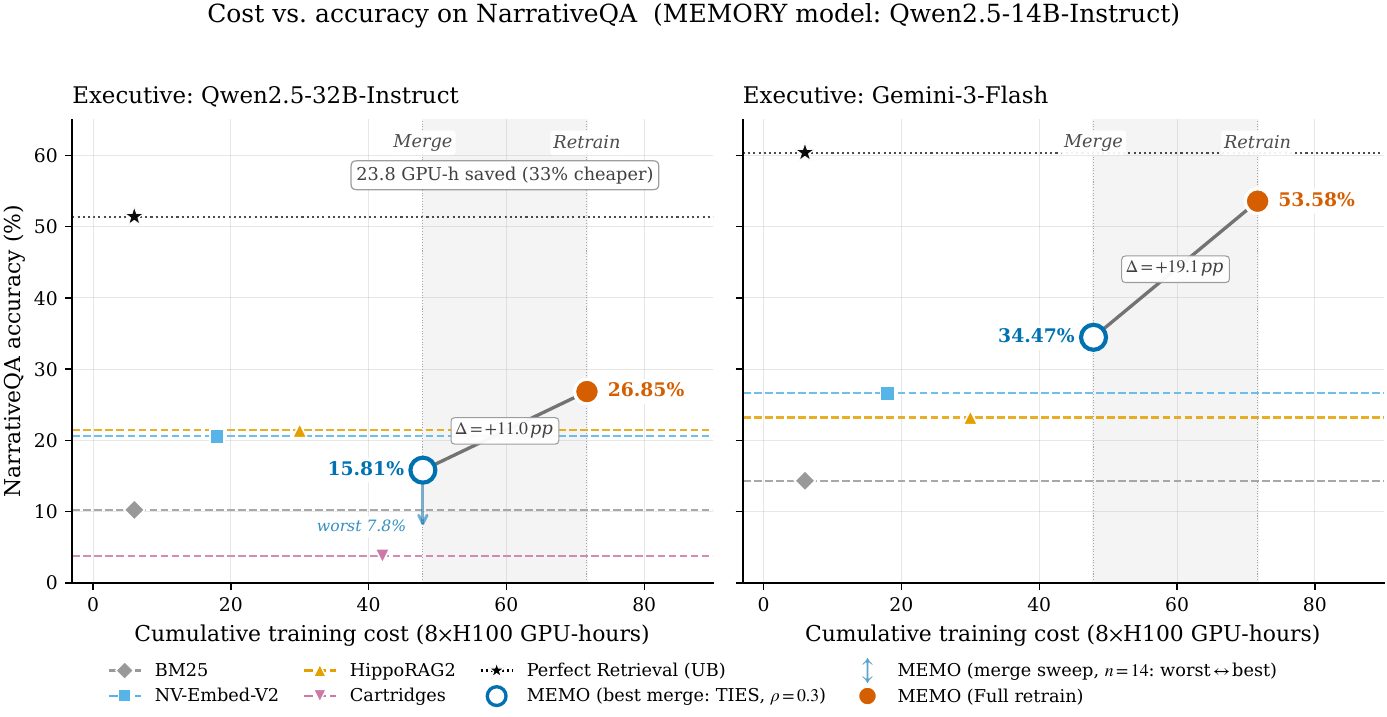}
    \caption{
        \textbf{Cost--accuracy trade-off on NarrativeQA when a second corpus arrives ($K{=}2$, \mem{} = Qwen2.5-14B-Instruct, 8$\times$H100).}
        Cumulative training cost is shown on the $x$-axis (one Qwen-14B SFT run takes $\approx 24$ GPU-hours on a 640k-QA-pair corpus). Merging trains \mem{} only on the new corpus, costing $X{+}Y \approx 48$~GPU-hours, while full retraining re-runs on the union, costing $X{+}(X{+}Y) \approx 72$~GPU-hours --- a 33\% saving. Merge-TIES ($\rho{=}0.3$) trails full retraining by $11.0$\% with Qwen2.5-32B-Instruct and $19.1$\% with Gemini-3-Flash as \main{}, but still outperforms all retrieval baselines (BM25, NV-Embed-V2, HippoRAG2, Cartridges). The vertical $\updownarrow$ at the merge cost shows the worst-to-best range across the 14 merge configurations swept (\cref{tab:merge-method-sweep}). Perfect Retrieval is shown as the upper bound.
    }
    \label{fig:cost_vs_accuracy}
    \vspace{-2mm}
\end{figure}

\para{Results.}
A single SFT run consumes $\approx 24$ GPU-hours on 8$\times$H100; after two arrivals, full retraining incurs $X{+}(X{+}Y){=}72$ GPU-hours of cumulative compute, whereas merging accumulates only $X{+}Y{=}48$ GPU-hours --- a 33\% reduction (\cref{fig:cost_vs_accuracy}). The asymptotic gap widens with $K$: under the same per-corpus cost, merging scales as $\Theta(K)$ while full retraining scales as $\Theta(K^2)$, yielding a $5.5{\times}$ saving at $K{=}10$ ($240$ vs.\ $1{,}320$ GPU-hours). On accuracy, Merge-TIES ($\rho{=}0.3$) trails full retraining by $11.0$\% with Qwen2.5-32B-Instruct as \main{} ($15.81\%$ vs.\ $26.85\%$) and by $19.1$\% with Gemini-3-Flash ($34.47\%$ vs.\ $53.58\%$), placing the merged \mem{} below the union-retrained \mem{} but above every retrieval baseline. The full per-method sweep is reported in \cref{tab:merge-method-sweep}: TIES ($\rho{=}0.3$) and DARE-Linear ($\rho{=}0.3$) lead at $15.81\%$ and $15.47\%$ respectively, while SLERP ($t{=}0.5$) is the worst configuration at $7.85\%$. The pattern across families suggests that aggressive sparsification at low $\rho$ paired with sign-conflict resolution (TIES, DARE-Linear) is the most reliable merging recipe in this regime. These results confirm the predicted compute--accuracy trade-off: merging recovers most of \mem{}'s headroom over retrieval methods at substantially lower cumulative cost.

\begin{table}[!ht]
    \centering
    \caption{
        \textbf{Sweep of all 14 merge configurations on NarrativeQA.}
        Two \mem{}s (Qwen2.5-14B-Instruct) are independently SFT-trained on the disjoint NarrativeQA.1 and NarrativeQA.2 subsets; each subset's best-performing checkpoint provides the task vector entering the merge. \main{} = Qwen2.5-32B-Instruct; results are mean $\pm$ std.\ dev.\ over 3 runs. Best merge in \textbf{bold}; full-retrain accuracy ($26.85 \pm 0.39$) is shown for reference.
        Hyperparameter conventions: $t \in [0,1]$ is the SLERP interpolation factor along the unit sphere connecting the two task vectors ($t{=}0$ recovers \mem{} on NarrativeQA.1, $t{=}1$ recovers \mem{} on NarrativeQA.2, $t{=}0.5$ is the geodesic midpoint); $\rho \in (0,1]$ is the sparsification density --- the fraction of largest-magnitude task-vector entries kept (TIES) or the keep probability for random-drop sparsification (DARE, DARE-TIES). Linear and Task arithmetic merge with uniform weights ($\lambda_i = 1$) and have no hyperparameter.
    }
    \label{tab:merge-method-sweep}
    \small
    \setlength{\tabcolsep}{8pt}
    \renewcommand{\arraystretch}{1.25}
    \begin{tabular}{|l|c|c|}
        \hline
        \textbf{Method family} & \textbf{Hyperparameter} & \textbf{Accuracy (\%)} \\
        \hline
        Linear           & ---           & $11.60 \pm 1.02$ \\
        Task arithmetic  & ---           & $12.74 \pm 1.75$ \\
        \hline
        \multirow{3}{*}{SLERP}        & $t=0.3$    & $11.60 \pm 2.24$ \\
                                      & $t=0.5$    & $\phantom{0}7.85 \pm 1.71$ \\
                                      & $t=0.7$    & $11.60 \pm 2.13$ \\
        \hline
        \multirow{3}{*}{TIES}         & $\rho=0.3$ & $\mathbf{15.81 \pm 0.39}$ \\
                                      & $\rho=0.5$ & $12.17 \pm 1.94$ \\
                                      & $\rho=0.7$ & $12.06 \pm 2.58$ \\
        \hline
        \multirow{3}{*}{DARE-Linear}  & $\rho=0.3$ & $15.47 \pm 0.79$ \\
                                      & $\rho=0.5$ & $\phantom{0}9.78 \pm 1.20$ \\
                                      & $\rho=0.7$ & $13.65 \pm 2.08$ \\
        \hline
        \multirow{3}{*}{DARE-TIES}    & $\rho=0.3$ & $11.72 \pm 0.52$ \\
                                      & $\rho=0.5$ & $12.97 \pm 1.23$ \\
                                      & $\rho=0.7$ & $11.04 \pm 1.20$ \\
        \hline
    \end{tabular}
\end{table}

\FloatBarrier

\section{Validating evaluation dataset suitability}
\label{app:EXPR-data-contamination}
\begin{table}[!ht]
    \vspace{-4mm}
    \centering
    \caption{Performance gap between no context and perfect retrieval across datasets and \main{}s.}
    \label{tab:data-contamination-results}
    \small
    \setlength{\tabcolsep}{6pt}
    \renewcommand{\arraystretch}{1.25}
    \resizebox{\columnwidth}{!}{%
    \begin{tabular}{|c|ccc|ccc|}
        \hline
        & \multicolumn{3}{c|}{\textbf{Qwen2.5-32B-Instruct}} 
        & \multicolumn{3}{c|}{\textbf{Gemini-3-Flash}} \\
        & \textbf{BrowseComp-Plus} & \textbf{NarrativeQA} & \textbf{MuSiQue}
        & \textbf{BrowseComp-Plus} & \textbf{NarrativeQA} & \textbf{MuSiQue} \\ \hline
        No Context & $0.00 \pm 0.00$ & $5.35 \pm 0.20$  & $17.03 \pm 0.40$ & $1.33$ & $26.62$ & $41.80$ \\ \hline
        Perfect Retrieval & $79.67 \pm 1.45$ & $51.42 \pm 0.52$  & $62.83 \pm 0.90$  & $88.33$ & $60.41$ & $73.00$ \\ \hline
    \end{tabular}%
    }
\end{table}
To assess the suitability of the evaluation datasets for \main{} and whether the \main{} has memorized answers from training data, we evaluate performance both without any context (\textbf{No Context}) and with evidence documents provided (\textbf{Perfect Retrieval}), the latter serving as an empirical upper-bound that assumes perfect retrieval of relevant documents.

As shown in \cref{tab:data-contamination-results}, the large disparity in performance between No Context and Perfect Retrieval confirms that these datasets require access to evidence documents to achieve correct answers, validating their suitability for evaluating \memo{}. 

Unsurprisingly, MuSiQue yields the highest No Context scores, as its Wikipedia-grounded questions fall within models' parametric knowledge. NarrativeQA proves most challenging as it achieves the lowest Perfect Retrieval scores across both \main{}s, reflecting the demand for careful reasoning over full-length books and movie scripts. BrowseComp-Plus yields the largest disparity between No Context and Perfect Retrieval, with near-zero No Context performance but strong recovery when evidence documents are provided.

These findings confirm that \main{} heavily relies on evidence documents across all three datasets to perform well. MuSiQue tests multi-hop factual reasoning where parametric knowledge provides partial signals, NarrativeQA tests narrative comprehension that remains challenging even with perfect context, and BrowseComp-Plus tests the ability to exploit retrieved documents for facts otherwise entirely inaccessible to the model.

\section{Evaluation details}
\label{app:EXPR-eval-process-and-prompt}

\subsection{Implementation details}
\label{subsec:eval-impl-details}
The current temperature settings are described in~\cref{tab:temperatures}. Stage 1 only has a budget of $1$ interaction, Stage 2 has a budget of $7$ interactions, Stage 3 has a budget of $8$ interactions.

\begin{table}[!ht]
    \vspace{-4mm}
    \centering
    \caption{Temperature Configuration of each Stage from~\cref{sec:inference}}
    \label{tab:temperatures}
    \small
    \setlength{\tabcolsep}{6pt}
    \renewcommand{\arraystretch}{1.25}
    \resizebox{\columnwidth}{!}{%
    \begin{tabular}{|l|c|c|p{5.5cm}|}
        \hline
        \textbf{Stage} & \textbf{Model} & \textbf{Temperature Value} & \textbf{Intent} \\
        \hline
        Evaluation Stage 1 -- Grounding & \main{} & 0.4 & Moderate exploration to generate diverse but focused sub-questions \\
        \hline
        Evaluation Stage 1 -- Grounding & \mem{} & 0.1 & Near-deterministic to ensure stable, consistent grounding answers \\
        \hline
        Evaluation Stage 2 -- Entity identification & \main{} & 0.4 & Moderate exploration to identify varied candidate entities without excess noise \\
        \hline
        Evaluation Stage 2 -- Entity identification & \mem{} & 0.1 & Near-deterministic to produce reliable entity-targeted answers \\
        \hline
        Evaluation Stage 3 -- Answer Seeking & \main{} & 1.0 & High exploration to maximally diversify sub-questions once the entity is confirmed \\
        \hline
        Evaluation Stage 3 -- Answer Seeking & \mem{} & 0.3 & Slightly relaxed determinism to allow nuanced answers while remaining consistent \\
        \hline
        Final Synthesis & \main{} & 0.3 & Low temperature to produce a consistent final answer \\
        \hline
    \end{tabular}%
    }
\end{table}

\begin{table}[!ht]
    \vspace{-4mm}
    \centering
    \caption{Helper Functions for Stage 2 and 3 of the Evaluation Pipeline}
    \label{tab:eval-entity-helpers}
    \small
    \setlength{\tabcolsep}{6pt}
    \renewcommand{\arraystretch}{1.25}
    \resizebox{\columnwidth}{!}{%
    \begin{tabular}{|l|c|p{9cm}|}
        \hline
        \textbf{Function} & \textbf{Stage} & \textbf{Intent} \\
        \hline
        Track uncertain answer streaks & Stage 2 & Maintains a running tally of how many unanswerable questions each candidate entity has accumulated across Stage~2, allowing the \main{} to progressively prioritize candidates that the \mem{} consistently cannot corroborate \\
        \hline
        Select the best candidate & Stage 2 & Fallback bridge from Stage~2 to Stage~3 when entity pinning ends without a confirmed entity. Selects the highest \main{}-ranked candidate, with ties broken by the order in which the \mem{} produced the candidates \\
        \hline
        Entity pivot correction & Stage 3 & Allows the pipeline to self-correct mid Stage~3 if the Stage~2 entity proves incorrect. When the \main{} nominates a different entity, the confirmed entity is overwritten and marked as unconfirmed so subsequent turns are aware it was not pinned through the full Stage~2 process \\
        \hline
    \end{tabular}%
    }
\end{table}

Beyond what is described in~\cref{sec:inference}, there are additional helper functions that help manage failure modes across Stage~2 and Stage~3. Within Stage~2, the uncertain answer streak tracker is called at the start of every entity-pinning interaction and its output is passed directly into the entity-pinning prompt, giving the \main{} a live view of which candidates the \mem{} has repeatedly failed to corroborate. This allows the \main{} to continuously re-rank and prune the candidate pool as evidence accumulates. When Stage~2 concludes without a confirmed entity, either because the \main{} explicitly exhausts its options or the interaction budget is reached, the best candidate selector acts as the bridge into Stage~3 by returning the top-ranked candidate. In cases where multiple candidates share the highest rank, the first candidate in the order produced by \main{} is selected. In both cases, the downstream Stage~3 prompt is informed of whether the entity was formally confirmed or merely a best guess. Finally, if Stage~3 reveals that the Stage~2 entity was incorrect due to persistent \mem{} failures, the entity pivot mechanism allows the \main{} to nominate a replacement entity mid-stage. The confirmed entity is then overwritten and marked as unconfirmed, ensuring subsequent stages treat it with appropriate uncertainty rather than the confidence of a fully pinned entity. 

\subsection{Ablations on evaluation setup}
\label{app:eval-setup-ablation}
To justify our structured multi-turn evaluation design, we compare against two baselines: a single-turn setup and an \emph{unstructured} multi-turn setup; in both cases, the same trained \mem{} is used and \main{} is held fixed. Results are reported in~\cref{tab:eval-setup-ablation}.

\begin{table}[!ht]
    \vspace{-3mm}
    \centering
    \caption{\memo{} accuracy results with Qwen2.5-32B-Instruct as \main{} and Qwen2.5-14B-Instruct as \mem{} across evaluation setups. The best performing epoch was used in comparison across all 3 setups, with mean $\pm$ std. dev. reported across 3 runs. Bold results indicate best performing results in the column.}
    \label{tab:eval-setup-ablation}
    \small
    \setlength{\tabcolsep}{6pt}
    \renewcommand{\arraystretch}{1.25}
    \resizebox{\columnwidth}{!}{%
    \begin{tabular}{|p{5cm}|c|c|c|}
        \hline
        \textbf{Evaluation Setup} & \textbf{BrowseComp-Plus Accuracy} & \textbf{NarrativeQA Accuracy} & \textbf{MuSiQue Accuracy} \\
        \hline
        Single turn evaluation & $32.56 \pm 1.58$ & $24.80 \pm 0.20$ & $37.57 \pm 1.15$ \\
        \hline
        \makecell[l]{Unstructured multi-turn evaluation \\ (15 turns)} & $47.33 \pm 0.88$ & $26.73 \pm 2.17$ & $40.13 \pm 1.12$ \\
        \hline
        \makecell[l]{Unstructured multi-turn evaluation \\ (50 turns)} & $48.67 \pm 1.00$ & $27.19 \pm 0.71$ & $40.57 \pm 0.31$ \\
        \hline
        \makecell[l]{Structured multi-turn evaluation \\ (7 Entity Identification turns + \\ 8 Answer seeking turns)} & $\mathbf{54.22 \pm 0.84}$ & $26.39 \pm 1.75$ & $\mathbf{48.30 \pm 1.25}$ \\
        \hline
        \makecell[l]{Structured multi-turn evaluation \\ (7 Entity Identification turns + \\ 15 Answer seeking turns)} & $51.44 \pm 2.41$ & $\mathbf{27.76 \pm 0.20}$ & $47.57 \pm 0.95$\\
        \hline
    \end{tabular}%
    }
\end{table}

In a single-turn interaction, \main{} first determines whether the question requires external memory retrieval, and if so, decomposes it into a set of sub-questions (Stage~1, \cref{sec:inference}) and poses them all simultaneously to \mem{}. \mem{} responds to each sub-question independently, and responses indicating uncertainty are discarded before the remaining answers are passed to \main{} for final synthesis. This design requires \main{} to commit to its full sub-question set before observing any responses, preventing it from reformulating uninformative queries, following up on answers that introduce new candidate entities, or correcting retrievals that are incomplete, contradictory, or anchored to the wrong entity. This is a fundamental limitation that is reflected in its consistently lowest performance across all three datasets (\cref{tab:eval-setup-ablation}).

A natural extension of the single-turn setting is an unstructured multi-turn interaction, where \main{} examines the responses from \mem{} and decides whether sufficient information has been gathered, or whether additional retrieval rounds are needed (Stage~3, \cref{sec:inference}). In this setting, \main{} is presented with the full history of question-answer pairs and prompted to either synthesize a final answer or generate a new batch of sub-questions targeting remaining gaps, repeating for up to $T$ interactions. While iterative retrieval yields clear improvements over the single-turn baseline, performance plateaus quickly when increasing from 15 to 50 interactions ($47.33 \pm 0.88$ to $48.67 \pm 1.00$ on BrowseComp-Plus, $26.73 \pm 2.17$ to $27.19 \pm 0.71$ on NarrativeQA, and $40.13 \pm 1.12$ to $40.57 \pm 0.31$ on MuSiQue), suggesting that iterative retrieval alone is insufficient.

The structured multi-turn setup (see~\cref{sec:inference,subsec:eval-impl-details}) outperforms the unstructured multi-turn baseline, with 8 answer-seeking interactions achieving the strongest overall performance for BrowseComp-Plus and MuSiQue. This is consistent with the expectation that explicit entity identification is well-suited to the multi-hop reasoning demands of these datasets. NarrativeQA, which tests discourse understanding over long documents, has the unstructured 15- and 50-interaction baselines ($26.73 \pm 2.17$ and $27.19 \pm 0.71$) initially outperform the structured setup with 8 answer-seeking interactions ($26.39 \pm 1.75$). Inference logs indicate that \main{} rarely utilizes the entity identification stage on NarrativeQA, likely because its questions are less reliant on resolving specific entities. 
Consequently, the fixed-entity identification budget effectively reduces the number of available answer-seeking interactions compared to unstructured baselines. Increasing the answer-seeking budget to 15 interactions recovers this gap, with NarrativeQA reaching $27.76 \pm 0.20$, surpassing both unstructured baselines. We hypothesize that this could be due to additional answer-seeking interactions continuing to surface useful signals without the risk of entity drift or state corruption that compound in open-domain multi-hop settings.

Unlike NarrativeQA, structured entity identification and state tracking in BrowseComp-Plus and MuSiQue introduce sensitivity to error accumulation as the number of answer-seeking interactions increases beyond the optimal budget. Additional interactions increase the risk of erroneous \mem{} responses corrupting the known facts state, and provide more opportunities for \main{} to commit to an incorrect intermediate entity via the entity pivot correction helper (\cref{tab:eval-entity-helpers}). Furthermore, more interactions dilute the correct signal with potentially incorrect answers at the final synthesis stage. These failure modes are partly a function of the reasoning capability of \main{}, as structured state maintenance demands strong in-context reasoning to accurately track entities and avoid premature entity commitment. Corroborating this, we observe in \cref{tab:main-results} that a stronger reasoning model as \main{} yields improved performance when paired with the same \mem{}, suggesting that these failure modes can be mitigated by scaling the reasoning capability of \main{}. 

The stage budget used in our experiments was selected without systematic tuning, and alternative settings may yield similar performance with greater token efficiency. We therefore leave a systematic study of the optimal interaction budget and \main{} selection as future work.

\section{Discussion on number of training epochs}
\label{app:EXPR-num-trng-epoch}
\begin{figure}[!ht] 
    \vspace{-2mm}
    \centering
    \includegraphics[width=0.7\linewidth]{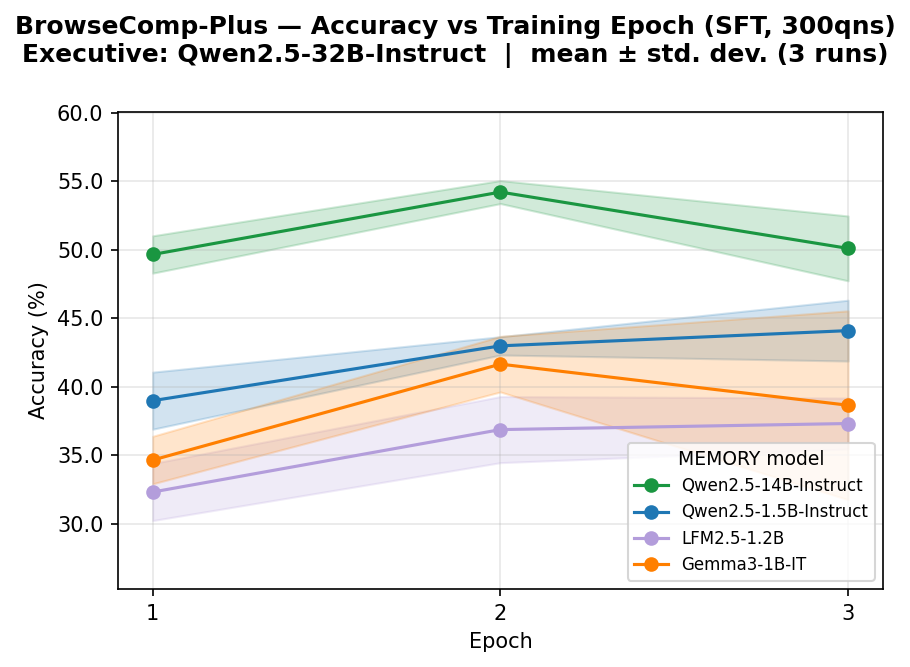}
    \caption{
        BrowseComp-Plus accuracy (\%) vs. training epoch (Full SFT) for each \memo{} model size and model family. Lines show the mean over 3 runs, and the shaded band shows $\pm$ std. dev. for Qwen2.5-32B-Instruct Runs. 
    }
    \label{fig:BCP-epoch-vs-acc}
    \vspace{-3mm}
\end{figure}

\begin{figure}[!ht] 
    \vspace{-2mm}
    \centering
    \includegraphics[width=0.7\linewidth]{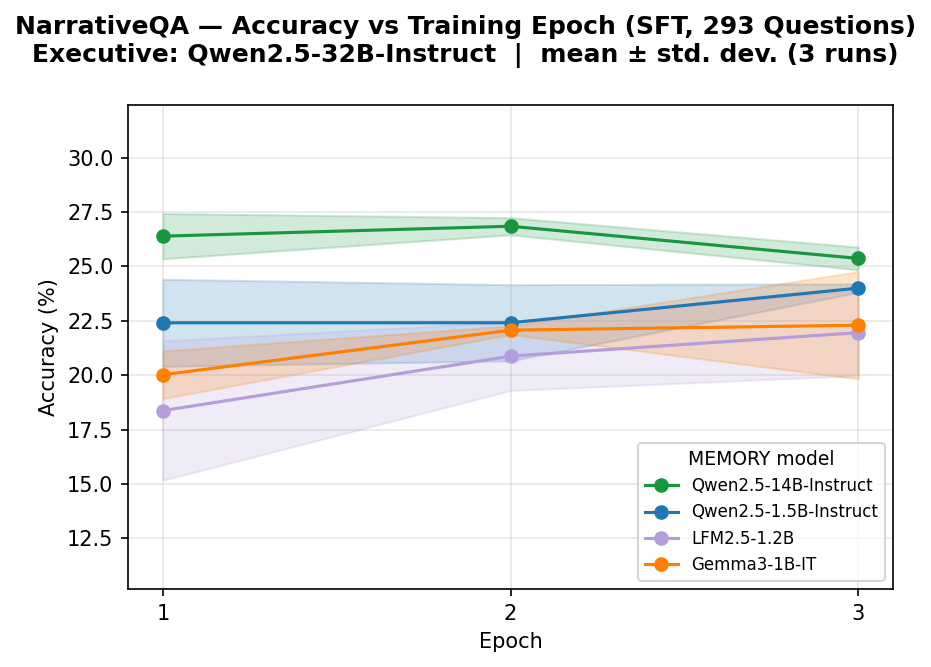}
    \caption{
        NarrativeQA accuracy (\%) vs. training epoch (Full SFT) for each \memo{} model size and model family. Lines show the mean over 3 runs, and the shaded band shows $\pm$ std. dev. for Qwen2.5-32B-Instruct Runs. 
    }
    \label{fig:NQA-epoch-vs-acc}
    \vspace{-3mm}
\end{figure}

\begin{figure}[!ht] 
    \vspace{-2mm}
    \centering
    \includegraphics[width=0.7\linewidth]{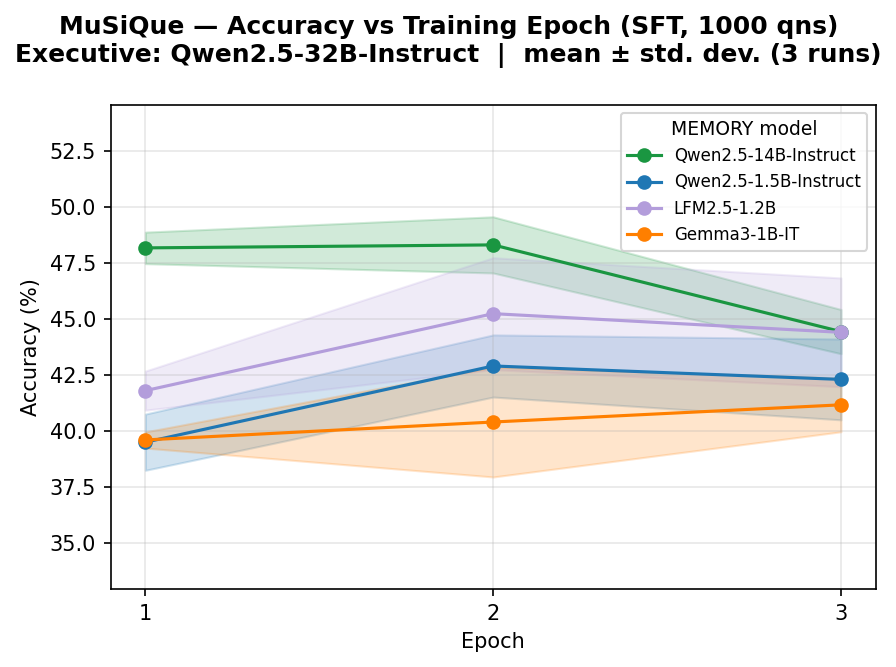}
    \caption{
        MuSiQue accuracy (\%) vs. training epoch (Full SFT) for each \memo{} model size and model family. Lines show the mean over 3 runs, and the shaded band shows $\pm$ std. dev. for Qwen2.5-32B-Instruct Runs. 
    }
    \label{fig:MSQ-epoch-vs-acc}
    \vspace{-3mm}
\end{figure}

From \cref{fig:BCP-epoch-vs-acc,fig:NQA-epoch-vs-acc,fig:MSQ-epoch-vs-acc}, we observe that additional training epochs do not consistently improve accuracy, as peak performance for most \mem{} occurs at epoch 2 with marginal gains or mild regression thereafter.
We attribute the early saturation and subsequent regression to overfitting on the SFT corpus, which exhibits substantial lexical overlap across steps by design, as later steps are derived from earlier ones~\cref{alg:datagen-pipeline}. To quantify this lexical overlap, we compute the lossless compression ratio of the combined QA text across all steps for each dataset by extracting all question and answer strings, concatenating them into a single text corpus, and applying gzip compression at maximum level (compression level 9), where the compression ratio is defined as the ratio of the original text size to the compressed size.

 BrowseComp-Plus ({$1{,}639{,}995$} pairs) achieves a ratio of {$5.80\times$} ({$82.8\%$} savings), MuSiQue ({$664{,}762$} pairs) achieves {$7.03\times$} ({$85.8\%$} savings), and NarrativeQA ({$1{,}276{,}676$} pairs) achieves {$5.45\times$} ({$81.7\%$} savings), indicating substantial lexical overlap within each dataset. We note that compression ratio captures lexical overlap only, and semantic diversity across QA pairs may remain higher, as each step targets distinct reasoning operations ranging from direct fact extraction to cross-document synthesis (see~\cref{alg:datagen-pipeline}), which is consistent with the impact of removing Step~5 (see~\cref{app:datagen-pipeline-Ablation}).

\section{Performance degradation of retrieval-based methods with increasing noise}
\label{app:perf-degradation-with-noise}
\begin{table}[!ht]
    \centering
    \caption{Accuracy (\%) on BrowseComp-Plus and MuSiQue with Qwen2.5-32B-Instruct as \main{}. \memo{} results are based on Qwen2.5-14B-Instruct and reported at the best training epoch. $N$ is the number of evidence documents present in the target corpus. $\Delta$ denotes accuracy difference (\%) compared to $0N$.}
    \label{tab:ablation-noise-robustness1}
    \small
    \setlength{\tabcolsep}{6pt}
    \renewcommand{\arraystretch}{1.25}
    \resizebox{\textwidth}{!}{%
    \begin{tabular}{|ll|c|cc|cc|}
        \hline
        \textbf{Method} & \textbf{Dataset} & \textbf{$0N$} & \multicolumn{2}{c|}{\textbf{$1N$}} & \multicolumn{2}{c|}{\textbf{$2N$}} \\
        & & Acc.\ (\%) & Acc.\ (\%) & $\Delta$ & Acc.\ (\%) & $\Delta$ \\
        \hline
        \multirow{2}{*}{NV-Embed-V2} & BrowseComp-Plus & $56.89 \pm 0.51$ & $50.67 \pm 0.33$ & ${\color{red}\downarrow 6.22}$ & $49.44 \pm 0.19$ & ${\color{red}\downarrow 7.45}$ \\
                                     & MuSiQue         & $42.30 \pm 0.53$ & $37.47 \pm 0.15$ & ${\color{red}\downarrow 4.83}$ & $33.03 \pm 1.10$ & ${\color{red}\downarrow 9.27}$ \\
        \hline
        \multirow{2}{*}{HippoRAG2}   & BrowseComp-Plus & $62.33 \pm 1.15$ & $56.11 \pm 0.51$ & ${\color{red}\downarrow 6.22}$ & $50.78 \pm 1.35$ & ${\color{red}\downarrow 11.55}$ \\
                                     & MuSiQue         & $47.33 \pm 0.74$ & $42.17 \pm 0.12$ & ${\color{red}\downarrow 5.16}$ & $41.70 \pm 0.69$ & ${\color{red}\downarrow 5.63}$ \\
        \hline
    \end{tabular}%
    }
\end{table}

\cref{tab:ablation-noise-robustness1} reports the performance of two retrieval-based baselines (NV-Embed-V2 and HippoRAG2) under increasing retrieval noise. Both methods degrade monotonically as noise increases, confirming their susceptibility to irrelevant documents. The degradation is most severe for HippoRAG2 on BrowseComp-Plus, which drops 11.55\% from $0N$ to $2N$, and for NV-Embed-V2 on MuSiQue, which drops 9.27\% over the same range. Notably, even a single negative document per evidence document ($1N$) causes substantial drops of up to 6.22\% for both methods on BrowseComp-Plus, suggesting that retrieval-based methods are extremely sensitive to noisy retrieval settings.

\section{Ablation on \mem{} size}
\label{app:ablate-mem-size}
Both Qwen2.5-1.5B-Instruct and Qwen2.5-14B-Instruct \memory{}s are trained on the same QA dataset generated by the \gen{} (Qwen2.5-32B-Instruct) under training settings described in~\cref{app:ALGO-train-settings}.
Each \mem{} is evaluated using Qwen2.5-32B-Instruct and Gemini-3-Flash as the \main{}.

\section{Ablation on \mem{} family}
\label{app:ablate-mem-family}
Each \mem{} is trained on the same QA dataset generated by \gen{} (Qwen2.5-32B-Instruct) and evaluated using Qwen2.5-32B-Instruct and Gemini-3-Flash as \main{}.
Notably, while Qwen2.5-1.5B-Instruct and Gemma3-1B-IT are based on standard transformer architectures, LFM2.5-1.2B-Instruct adopts a hybrid architecture combining state-space convolution with transformer attention blocks, thereby providing a broader test of \mem{} across diverse model designs. These models are trained on the same training settings in~\cref{app:ALGO-train-settings}, with Gemma3-1B-IT using eager attention during training instead of Flash Attention 2.

\section{Comparison between full SFT and LoRA}
\label{app:EXPR-sft-vs-lora}
We train all models using LoRA~\citep{hu2022lora} applied to the attention  and feed-forward projection layers: \texttt{q\_proj}, \texttt{k\_proj}, \texttt{v\_proj}, \texttt{o\_proj}, \texttt{gate\_proj}, \texttt{up\_proj}, and \texttt{down\_proj}. The general LoRA configuration is summarised in Table~\ref{tab:lora_training_config}, with model-specific rank and scaling settings reported in Table~\ref{tab:lora_models_config}. All remaining training hyperparameters follow Table~\ref{tab:training_config}, and per-dataset batch sizes are given in Table~\ref{tab:batch_sizes}.

\begin{table}[!ht]
    \centering
    \caption{LoRA Specific Training Configuration. All other parameters are the same as those in~\cref{tab:training_config}.}
    \label{tab:lora_training_config}
    \small
    \setlength{\tabcolsep}{6pt}
    \renewcommand{\arraystretch}{1.25}
    \begin{tabular}{|l|c|}
        \hline
        \textbf{Parameter} & \textbf{Value} \\
        \hline
        Target modules & \texttt{q\_proj, k\_proj, v\_proj, o\_proj,} \\
                       & \texttt{gate\_proj, up\_proj, down\_proj} \\
        \hline
        LoRA dropout & 0.05 \\
        \hline
        Bias & None \\
        \hline
        Learning rate & $2 \times 10^{-4}$ \\
        \hline
    \end{tabular}
\end{table}

\begin{table}[!ht]
    \centering
    \caption{Model-Specific LoRA Configuration.}
    \label{tab:lora_models_config}
    \small
    \setlength{\tabcolsep}{6pt}
    \renewcommand{\arraystretch}{1.25}
    \begin{tabular}{|l|c|c|c|c|}
        \hline
        \textbf{Model} & \textbf{Size} & \textbf{LoRA rank} & \textbf{LoRA alpha} & \textbf{Trainable params} \\
        \hline
        LFM2.5-1.2B-Instruct  & 1.2B &  8 & 16 & 6.1M (0.41\%) \\
        \hline
        Gemma3-1B-IT           & 1B   &  8 & 16 & 6.6M (0.65\%) \\
        \hline
        Qwen2.5-1.5B-Instruct  & 1.5B &  8 & 16 & 9.2M (0.60\%) \\
        \hline
        Qwen2.5-14B-Instruct   & 14B  & 16 & 32 & 68.8M (0.47\%) \\
        \hline
    \end{tabular}
\end{table}

\begin{table}[!ht]
    \centering
    \caption{Ablation on LoRA vs Full SFT training across all \mem{}s, evaluated with Qwen2.5-32B-Instruct as \main{}. All results are mean $\pm$ std. dev. over 3 runs. Bold results indicate best performing results in the column.}
    \label{tab:appendix-ablation-lora-sft}
    \small
    \setlength{\tabcolsep}{6pt}
    \renewcommand{\arraystretch}{1.25}
    \resizebox{\columnwidth}{!}{%
    \begin{tabular}{|c|cc|cc|cc|}
        \hline
        & \multicolumn{2}{c|}{\textbf{BrowseComp-Plus}} 
        & \multicolumn{2}{c|}{\textbf{NarrativeQA}} 
        & \multicolumn{2}{c|}{\textbf{MuSiQue}} \\
        \textbf{\mem{}} 
        & \textbf{LoRA} & \textbf{Full SFT}
        & \textbf{LoRA} & \textbf{Full SFT}
        & \textbf{LoRA} & \textbf{Full SFT} \\ \hline
        Gemma3-1B-IT           & $25.22 \pm 1.39$ & $41.67 \pm 2.03$ & $21.62 \pm 0.86$ & $22.30 \pm 2.47$ & $26.17 \pm 1.10$ & $41.17 \pm 1.20$ \\ \hline
        LFM2.5-1.2B-Instruct   & $0.78 \pm 0.19$ & $37.33 \pm 1.86$ & $5.69 \pm 0.71$ & $21.96 \pm 1.97$ & $7.50 \pm 0.26$ & $45.23 \pm 2.49$ \\ \hline
        Qwen2.5-1.5B-Instruct  & $29.78 \pm 0.51$ & $44.11 \pm 2.22$ & $21.84 \pm 0.34$ & $24.00 \pm 0.20$ & $31.53 \pm 0.55$ & $42.90 \pm 1.39$ \\ \hline
        Qwen2.5-14B-Instruct   & $\mathbf{48.78 \pm 1.02}$ & $\mathbf{54.22 \pm 0.84}$ & $\mathbf{23.78 \pm 0.52}$ & $\mathbf{26.85 \pm 0.39}$ & $\mathbf{43.94 \pm 0.97}$ & $\mathbf{50.07 \pm 0.81}$ \\ \hline
    \end{tabular}
    }
\end{table}

The notably poor LoRA performance of LFM2.5-1.2B-Instruct can be attributed to its hybrid convolution--attention architecture, which differs from the standard transformer models in our evaluation. Following the LFM2 architecture~\citep{amini2025lfm2}, LFM2.5-1.2B-Instruct consists of $16$ layers --- $6$ grouped-query attention (GQA) blocks (at indices $\{2,5,8,10,12,14\}$) interleaved with $10$ short-range LIV convolution (ShortConv) blocks~\citep{amini2025lfm2}. Crucially, the LFM2 attention output projection is named \texttt{out\_proj} (rather than \texttt{o\_proj}) and its SwiGLU MLP uses \texttt{w1}/\texttt{w3}/\texttt{w2} (rather than \texttt{gate\_proj}/\texttt{up\_proj}/\texttt{down\_proj}), while the ShortConv blocks expose their own \texttt{in\_proj} and \texttt{out\_proj} layers. A LoRA configuration targeting the standard Llama-family module names therefore adapts only a strict subset of the projections that exist in LFM2.5, leaving the remainder frozen. The result is $6.1$M trainable parameters ($0.41\%$ of total), disproportionately low given the model size and below our target of ${\sim}0.5\%$. The rank $r=8$ was kept fixed across all sub-2B models for a controlled comparison; in retrospect, this penalises LFM2.5-1.2B-Instruct due to its architectural mismatch with the standard Llama-style target set.

Furthermore, the $10$ ShortConv blocks which handle the bulk of the model's local feature extraction and the SwiGLU MLPs attached to every block remain entirely unadapted under standard LoRA targeting, severely limiting the adapter's ability to shift the model's behaviour. As shown in~\cref{tab:appendix-ablation-lora-sft}, the large performance gap between LoRA and Full SFT confirms that the model is capable of learning the task when all parameters are updated.

Future work could explore LoRA configurations better suited to this architecture: targeting the LFM2-specific module names (\texttt{out\_proj}, \texttt{w1}, \texttt{w3}, \texttt{w2}) alongside the ShortConv projections (\texttt{in\_proj}, \texttt{out\_proj}), as well as tuning the rank and learning rate per architecture rather than holding them fixed across families for the controlled comparison reported here.\newline